\pdfoutput=1
\documentclass[10pt,journal,compsoc]{IEEEtran}
\usepackage[font={footnotesize}]{caption}
\usepackage{subfig}

\ifCLASSOPTIONcompsoc
  \usepackage[nocompress]{cite}
\else
  \usepackage{cite}
\fi
\ifCLASSINFOpdf
\else
\fi

\usepackage[pdftex]{graphicx} \pdfcompresslevel=9

\usepackage{multirow}
\usepackage{sidecap}

\usepackage{amssymb}
\usepackage{amsmath}

\usepackage[hyphens]{url}
\usepackage[final]{changes}

\usepackage{color}
\definecolor{darkblue}{rgb}{0,0,0}
\newcommand\bk[1]{{\color{darkblue}#1}}
\definecolor{darkgreen}{rgb}{0.0,0.0,1}

\newcommand\ie

\newcommand{\argmin}{\operatornamewithlimits{argmin}}

\begin{document}

\title{Efficient Unsupervised Temporal Segmentation of Motion Data}
%
%
%
%

\author{Bj\"orn~Kr\"uger,~\IEEEmembership{}
        Anna~V\"ogele,~\IEEEmembership{}
        Tobias~Willig,~\IEEEmembership{}
        Angela~Yao,~\IEEEmembership{}\\
        Reinhard~Klein,~\IEEEmembership{Member,~IEEE}
        and~Andreas~Weber,~\IEEEmembership{Member,~IEEE}
\IEEEcompsocitemizethanks{
\IEEEcompsocthanksitem B. Kr\"uger is with Gokhale Method Institute, Stanford (CA), United States.\protect\\
E-mail: kruegerb@cs.uni-bonn.de

\IEEEcompsocthanksitem A. V\"ogele, T. Willig, A. Yao, R. Klein and A. Weber are with the Department
of Computer Science, University of Bonn, Germany.\protect\\
}

\thanks{Manuscript received April, 2015; revised October, 2015.}}

%
%

\markboth{Submitted to IEEE TPAMI,~Vol.~XX, No.~X, XXX~2015}%
{Shell \MakeLowercase{\textit{et al.}}: Motion Segmentation}
%



\IEEEtitleabstractindextext{%
\begin{abstract}
We introduce a method for automated temporal segmentation of human motion data into distinct actions and compositing motion primitives based on self-similar structures in the motion sequence. We use neighbourhood graphs for the partitioning and the similarity information in the graph is further exploited to cluster the motion primitives into larger entities of semantic significance. The method requires no assumptions about the motion sequences at hand and no user interaction is required for the segmentation or clustering. In addition, we introduce a feature bundling preprocessing technique to make the segmentation more robust to noise, as well as a notion of motion symmetry for more refined primitive detection. We test our method on several sensor modalities, including markered and markerless motion capture as well as on electromyograph and accelerometer recordings. The results highlight our system's capabilities for both segmentation and for analysis of the finer structures of motion data, all in a completely unsupervised manner.

\end{abstract}

}

\maketitle

\IEEEdisplaynontitleabstractindextext

%
\IEEEpeerreviewmaketitle

\ifCLASSOPTIONcompsoc
\IEEEraisesectionheading{\section{Introduction}\label{sec:introduction}}
\else
\section{Introduction}
\label{sec:introduction}
\fi

\IEEEPARstart{H}{uman motion} capture, once associated with producing special effects for films and video games, is \replaced{common today}{today commonplace} in \replaced{diverse}{a diverse array of} applications ranging from health care to consumer electronics. The ever-increasing simplicity to capture data by different sensor modalities, along with the sheer amount of existing recorded data creates a demand for motion analysis methods that are computationally efficient, \replaced{yet with minimal}{all the while minimizing the amount of required} human input.

Dividing streams of motion data into perceptually meaningful segments is a precursor to almost all analysis and synthesis methods. For example, creating a statistical motion model calls for data already preprocessed into well-defined activity segments.  Further segmentation of these activities into individual cycles is greatly beneficial for action recognition, especially when reptitions should be counted, or \replaced{for compressing}{the compression of} motion data. However, the quantity of captured data does not always allow for time-consuming manual segmentation.  As such, unsupervised segmentation and learning of motion primitives is a topic of interest that has been addressed in both the computer vision~\cite{YongRui2000,ZelnikManor2006,DelVecchio2003,Lu2004,Zhou08,Zhou13,Hoai2011,Araujo2014}
and the computer graphics~\cite{Barbic04,Mueller2005,Liu2006,voegele2014a,hou2015} communities. 


We propose a segmentation method which identifies \textbf{distinct actions} within motion sequences and further decompose such actions into atomic \textbf{motion primitives}, all in an unsupervised fashion. For example, in a sequence of a person who first walks and then breaks into a run, we can separate walking from running, as well as the individual steps of the walk and run.  We identify both the actions and the motion primitives by exploiting the self-similarities that exist in motion sequences.



We pose segmentation as an efficiently solvable graph problem, as first presented in~\cite{voegele2014a} for segmenting motion capture data. To further improve computational efficiency, we employ a Neighborhood Graph~\cite{Krueger10}. 
In addition, we propose three new contributions, making the method robust and applicable to motion data from varying sensor modalities.
First, we propose a novel feature bundling technique for preprocessing motion features.  The feature bundling allows for compact model representations of the motions, as well as robustness to noise, to accommodate modalities such as markerless motion capture or accelerometers. 
Second, we introduce a notion of motion symmetry, and exploit this as a means of refining primitive detection.  Considering symmetry often leads to primitives with more physical meaning---for instance walking cycles can be split into left and right steps. 
Third, we propose a clustering method for the detected motion segments also based on the self similarities which needs no assumption on the number of clusters.
We show the applicability of our method on a wide variety of motion datasets, ranging from markered and markerless motion capture to accelerometer and surface electromyography (EMG) recordings.

Defining the segments based on self-similarity gives our method several advantages over previous segmentation methods. First, it allows us to distinguish not only the distinct action segments, but also the transition segments between actions. Explicit treatment of transitions has so far not been addressed in previous works on unsupervised segmentation~\cite{Zhou08,Zhou13,MinChai2012}.  However, it has a direct impact on synthesis methods, such as motion graphs~\cite{KovarGleicher2002,MinChai2012}; not handling the transitions forces synthesized sequences to include additional primitives coming from possibly unrelated transitions, although this is neither convenient nor intrinsically motivated.

The second advantage of our method is that resulting motion primitives are highly consistent, with similar start and end points from sequence to sequence. Previous approaches often yield primitives which are phase-shifted from one another~\cite{Zhou13}. Any statistical \replaced{model}{models} built on unaligned primitives will be noisier and less representative of the actual motion.  We note \deleted{however}, that our motion primitives are not limited to being full motion cycles i.e.\ those starting and ending in the same body pose, but can also be either parts of a cycle or entirely non-cyclic in nature.

The rest of this paper is organized as follows. First, the feature bundling technique is introduced in Section~\ref{bundling}. Details of the motion segmentation into actions and subsequent motion primitives are described in Section~\ref{sec:segmentation}. In Section~\ref{motion_symmetry}, we introduce a notion of symmetry to help refine the segmented motion primitives.
In Section~\ref{sec_clustering}, the clustering of the motion primitives is discussed. We present segmentation results from motion sequences of varying modalities in Section~\ref{sec:experiments}, and conclude by discussing the achievements and limitations of our novel method as well as possible extensions in Section~\ref{sec:limitations}. Source code for the algorithm is available online
upon acceptance of this work.

\section{Related Work}

Temporal segmentation is related to a number of different fields such as data mining~\cite{Keogh2001,Fearnhead2006}, audio and speech processing~\cite{Ostendorf96,MuellerGW09_FolkSongAnnotation_ISMIR,MuellerG12_FolkSongSegmentation_ITG,PraetzlichM14_AudioTrackSeg_ISMIR}, and behavioural pattern recognition~\cite{Xuan2007}.
In statistical terms, the segmentation problem can be posed as a change point detection task~\cite{Fearnhead2006} which has been extended to a multi-dimensional setting~\cite{Xuan2007} based on the established Bayesian techniques~\cite{NIPS2009_3685}.
From a more general signal processing point of view, kernel-based methods for change-point analysis~\cite{Harchaoui2008,Desobry2005},
Hidden Markov Models as means for training and recognition~\cite{Ostendorf96},
and audio thumbnailing combined with enhanced similarity matrices~\cite{MuellerGW09_FolkSongAnnotation_ISMIR,MuellerG12_FolkSongSegmentation_ITG},
have been introduced.  There have been a number of attempts to solve the non-trivial problem of automatic segmentation of human motion data, which we outline below.

\subsubsection*{Pose Clustering}
One strategy for segmentation is based on clustering the poses present in a time series~\cite{Beaudoin2008,Gall12,bernard2013}. Beaudoin et al.~\cite{Beaudoin2008} proposed to extract motion motifs as building blocks of graphs.
Gall et al.~\cite{Gall12} create temporally meaningful pose clusters associated with unlabeled actions.
Bernard et al.~\cite{bernard2013} proposed an exploratory search and analysis system called \textit{MotionExplorer} based on similarity features.
Depending on the aggregation level, shorter or longer motion segments were found which may be associated with isolated human actions. However, Bernard's main focus was on a new visual representation of motion data rather than analysis tasks as discussed here.

\subsubsection*{Examplars and Template Models}
An alternative approach for segmentation is to apply example segments or pre-computed templates and match them to test sequences.
\added{For example, Lv et al.~\cite{Lv2006} uses a combination of HMMs and AdaBoost to learn discriminative feature templates from labeled segments to perform action recognition and segmentation.
}  M{\"u}ller et al.~\cite{Mueller2005,Mueller2006,Mueller2009} developed a framework based on geometric features to learn templates for solving and accelerating solutions to matching problems such as annotation and retrieval~\cite{Mueller2006,Mueller2009}. Adaptive segmentation~\cite{Mueller2005} is a fundamental result from using geometric feature vector sequences to compare motion capture data streams at the segment level rather than frame level. Another set of approaches~\cite{LanSun2013,Salamah2015} present related ideas on learning intrinsic regularities for segmentation and demonstrate that motion capture data can be segmented by using only a limited set of example motions, even when belonging to different action types than that of the training data.  Template approaches work well if the templates are available; our work is targeted at cases in which the exemplars or templates are not known beforehand.

\subsubsection*{Motion Synthesis}
Segmentation has also been addressed in conjunction with motion synthesis, such as motion concatenation~\cite{KovarGleicher2002} and motion parameterization~\cite{KovarGleicher2004}.  In motion concatenation, a \textit{motion graph} is constructed from clips of motion capture data; new sequences are then sythesized by motion extraction on this graph. In motion parameterization, motion elements are retrieved from large \replaced{datasets}{datasetse}, based on similarity to a query motion, and then blended together according to user constraints. The novel distance relation used in this work~\cite{KovarGleicher2004} has become the standard for finding similar motion clips at interactive speeds.

Later works~\cite{HeckGleicher2007,SafonovaHodgins2007,MinChai2012} combine both these ideas to accomplish motion synthesis techniques for high quality interactive applications. Min and Chai's Motion Graphs++~\cite{MinChai2012} is an advanced combination which effectively enables a variety of applications such as motion segmentation, recognition and online synthesis. All these approaches have a need for meaningful motion primitives that can be clustered to build statistical motion models or at least to allow for interpolation. Typically these are found by manual selection of some example primitives and a retrieval component to search for further exemplars in a database.

\subsubsection*{Unsupervised Motion Segmentation}

The methods most similar to ours are \replaced{those}{thos} which segment motions in an unsupervised way. Partitioning motion sequences into behavior segments by a PCA-based method was proposed by Barbic et al.~\cite{Barbic04}. This segmentation focuses on detecting behavior segments and is similar in spirit to the first step of our approach for isolating distinct actions. In~\cite{Barbic04}, the quality of local PCA models is tracked temporally; new activities are defined when the old PCA model cannot capture the data variance and a new PCA model is required. This approach is neither able to separate activities that fit into one local model, nor is able to detect individual representations.

The groundbreaking work of Zhou et al.~\cite{Zhou08,Zhou13} uses (hierarchically) aligned cluster analysis (H)ACA to temporally cluster poses into motion primitives which are then assigned to different action classes.  These approaches use kernel based projections and a time alignment to compare motion primitives of varying length. An initial segmentation of uniform length is refined by the clustering approach though the final resulting segments do not vary much from the initial length. One of the major differences between the work of Zhou et al. and ours is that the transitions between distinct actions are not considered.  Transition frames are assigned either to the previous or the following action segment, thereby reducing the consistency of the primitives.

\added{Finally, Gong et al.~\cite{Gong14} present an online approach based on \emph{kernelized temporal cuts} which incorporate Hilbert space embedding of distributions when extending change-point detection methods. This work is not directly comparable to ours as it is an online approach.}

\section{Feature Bundling}\label{bundling}

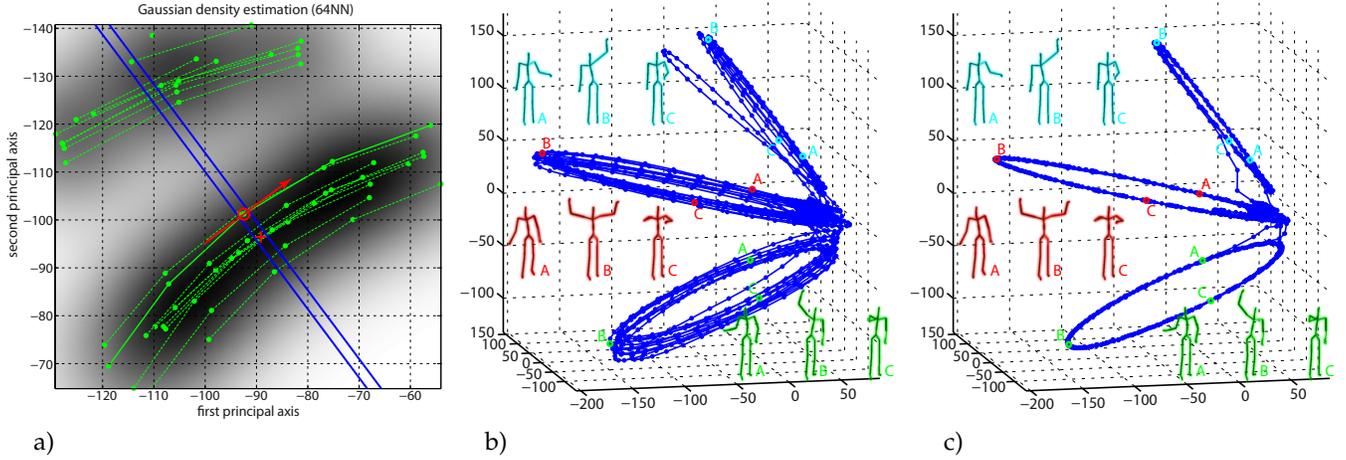
\begin{figure*}[t!]
  \centering
  \includegraphics[width=0.32\textwidth]{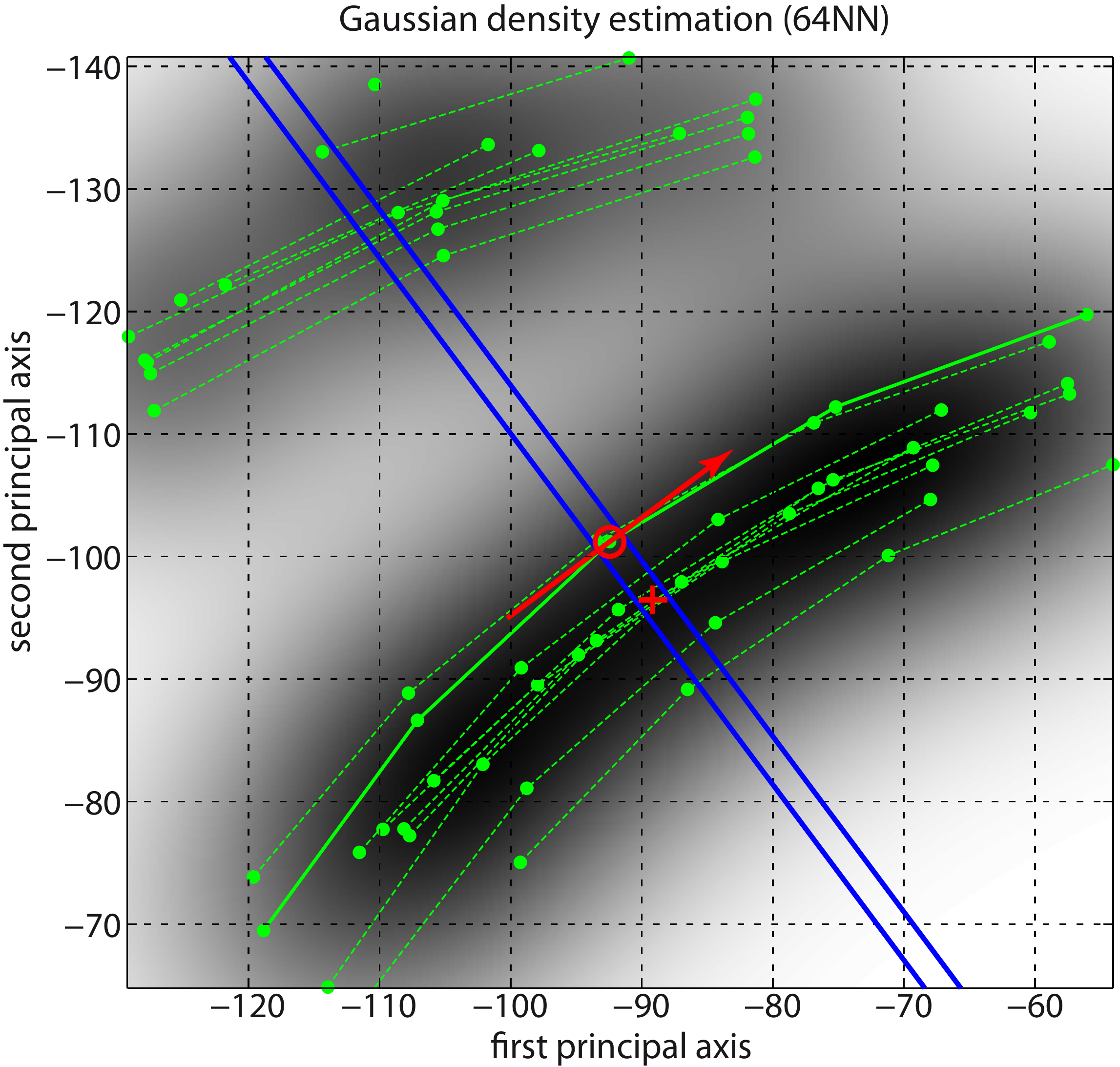}
  \includegraphics[width=0.65\textwidth]{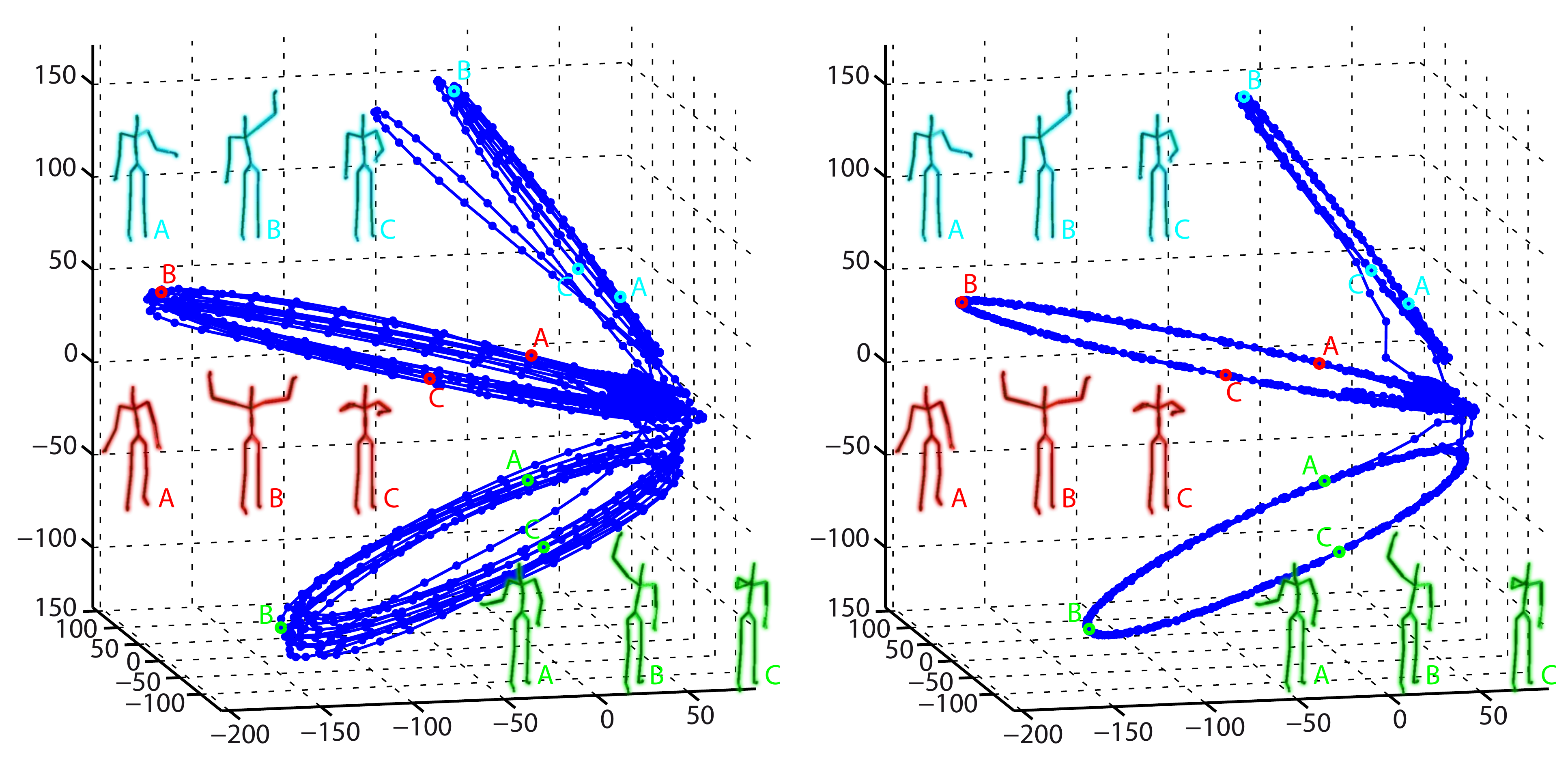}

    \setlength{\unitlength}{1cm}
    \begin{picture}(0,0)(0,0)
    	\put ( -8.5,0){\mbox{a)}}
    	\put ( -2.5,0){\mbox{b)}}
        \put (  3.6,0){\mbox{c)}}
    \end{picture}
  \caption{
  a) 2D example for \textbf{density estimation}: Original data points (red circle), 64 \textbf{nearest neighbors} (green dots connected in time), the \textbf{direction of movement} (red arrow), the 1D \textbf{subspace} used for optimization (blue lines) and the resulting position (red cross). b) 3D \textbf{PCA projection} of the feature sets of a mocap sequence (CMU 86\_11). The three sections of loops correspond to repetitions of differing arm rotation movements as indicated by representative body poses. c) \textbf{Bundled feature points} for the same motion capture sequence. Figure is best viewed in colour.}
  \label{fig:featurecondensation}
\end{figure*}

Semantically or visually similar motions, even when represented in dedicated feature spaces, may still differ notably due \added{to} variations in the performance of each action or cycles within the same action. Inspired by \added{the idea of} edge bundling \deleted{techniques} used in visualization~\cite{Holten2006,Bottger2014}, \added{where similar edges of a graph are visualized together for a better overview,} we propose a \replaced{bundling of similar}{similar bundling of} features. The goal of feature bundling is to topologically align disjoint motions; motions of same action class, but exhibiting differences in the feature space as a result of performance variations can be bundled together, while motions belonging to different classes or showing entirely different styles should be kept apart. \added{Note that even though we name our technique feature bundling, we have completely different objectives as well as different methodological approaches than \emph{bundle adjustments} used in 3D reconstruction algorithms.}



In our bundling technique we use a density estimation based optimization method which adjusts each point orthogonally to the direction of its trajectory in the feature space, effectively projecting the features onto a smoother manifold (see Figure~\ref{fig:featurecondensation} for an overview). Vejdemo-Johansson et al.~\cite{vejdemo2014a} consider a related idea by computing a typical motion cycle of a set of similar periodic motions.

For each frame $i$ of an input sequence of length $N$, a feature vector $F_i$ where $i\in [1\hdots N]$ is computed; the specific feature depends on the sensor modality (more details in Section~\ref{sec:experiments}). The goal is to compute a new feature vector $\hat{F}_i$ representative of $F_i$, but closer to other features at the same stage of the same motion cycle. Specifically,

\begin{enumerate}
\item\label{it:knn} For each $F_i$, find \textbf{$k$ nearest neighbors} within all other features $F_j, \ j \in [1\hdots N]\setminus i$.
\item\label{it:dirmovement}\label{it:subspace} Compute \textbf{direction of movement $d_i$} for $F_i$ and build a $D\!\!-\!\!1$ dimensional \textbf{subspace $\mathcal{S}_i$}  orthogonal to $d_i$.
\item\label{it:opt} \textbf{Optimize} $\hat{F}_i$ using a \textbf{kernel density estimate} of $F_i$ in $\mathcal{S}_i$ based on the $k$ nearest neighbors.
\item\label{it:backproj} Backproject $\hat{F}_i$ to the original feature space for the final representation.
\end{enumerate}

For the \textbf{$\mathbf{k}$-nn search}
we use a kd-tree as per~\cite{Krueger10} to find similar frames within the motion sequence.
We do not use a fixed search radius since the distance between sample points may vary to a great extent. Instead, a fixed number of k nearest neighbors ensures that the model reflects the local density of samples.

The \textbf{direction of movement} $d_i$
for frame $i$ is given by the numerically centered five-point derivative at this frame. Constructing an orthogonal \textbf{subspace}
prevents the overall data structure from collapsing. The basis of this subspace is computed via QR decomposition of a $D\!\!\times\!\!D$ matrix, where the first column vector is set to the direction of movement, while all other entries are filled with random numbers.

The local kernel \textbf{density estimation}
is based on the $k$ nearest neighbors and characterizes the data distribution of $F_i$ with kernel function $K_H$, i.e.\ a symmetric multivariate density with bandwidth matrix $H$:
\begin{equation}
K_{H^i}(x)=|H^i|^{-\frac{1}{2}}K({H^i}^{-\frac{1}{2}}x).
\end{equation}
Since our data is assumed to be Gaussian, we use a general approximation of the bandwidth matrix which minimizes the MISE (refer to Scott~\cite{Scott92}, Chapter $6$, Scott's rule) by
\begin{equation}
H_{jj} = \sigma_j k^{\frac{-1}{d+4}},
\end{equation}
\noindent where $j=1,\ldots, d$ and $\sigma_j$ is the standard deviation of the $j$th variate.
We use the multivariate Gaussian kernel:
\begin{equation}
K(x,\mu,\sigma) = e^{-\frac{1}{2}(x-\mu)'\Sigma^{-1}(x-\mu)},
\end{equation}
with $x=\left(x_1,\ldots, x_d\right)$, $\mu=\left(\mu_1,\ldots,\mu_d\right)$ the vector of empirical mean values, and $\Sigma$
the sample covariance matrix.

Note that the kernel density estimation is compatible with a preceding dimensionality reduction step. In fact, it is possible to reliably estimate the kernel density function without increasing the sample size $k$~\cite{Scott92}. Therefore, we apply a PCA to each set of samples in advance to reduce the dimension while keeping $97.5\%$ of the sample set's variance.  In principle, any of the commonly used dimensionality reduction techniques is applicable; we prefer PCA for its simplicity and speed, though other approaches have been explored extensively in the past~\cite{Jenkins2004,Barbic04,Filho2007}.

The \textbf{optimization}
searches for the new position $\hat{F}_i$ and is posed as an energy minimization problem. The offset $O_i$ from $F_i$ is optimized with respect to the density estimation, resulting in the final feature position $\hat{F}_i$:
\begin{equation}
\hat{F}_i = \argmin_{O_i} K_{H^i}(F+O_i).
\end{equation}

\begin{figure*}[t]
  \centering
  \includegraphics[width=0.9\linewidth]{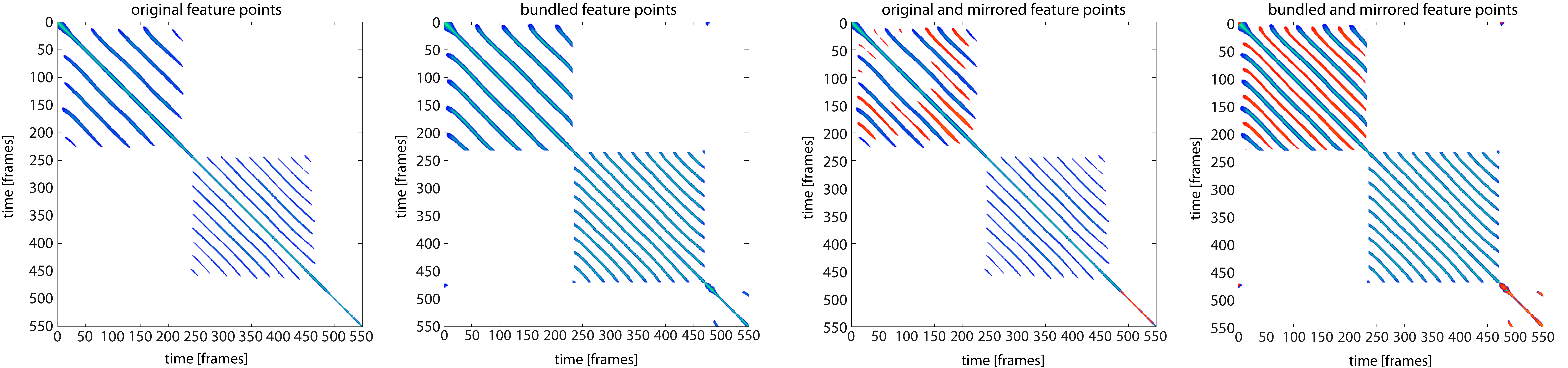}
  \setlength{\unitlength}{1cm}
    \begin{picture}(0,0)(8,0)
    	\put ( -8.6,-0.1){\mbox{a)}}
    	\put ( -4.3,-0.1){\mbox{b)}}
    	\put ( -0.1,-0.1){\mbox{c)}}
    	\put (  3.8,-0.1){\mbox{d)}}
    \end{picture}
  \caption{Comparison of SSSMs of two different situations: original and bundled features. Part a) shows the SSSM of original features as they were extracted from the input motion. Part b) shows the same SSSMs bur of the bundled features. Images c) and d) include the mirrored features also (highlighted red). Note that the SSSM based on the original mirrored features, the structure in the SSSM (c)) is less consistent than in SSSM of the bundled representation (d)).}
  \label{fig:featurecondensation_sssm}
\end{figure*}
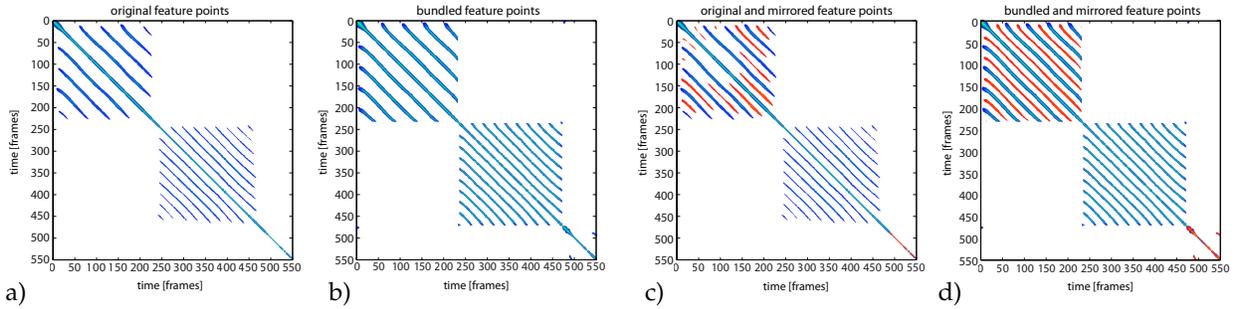

An example of the feature bundling is given in Figure~\ref{fig:featurecondensation}~a). A 3D projection of input and output feature points of a motion sequence are given in Figure~\ref{fig:featurecondensation}~b) and c), respectively.  A comparison between a sparse self similarity matrix (SSSM) based on the original and the bundled features is given in Figure~\ref{fig:featurecondensation_sssm} a) and b) respectively.

\section{Segmentation Technique}\label{sec:segmentation}

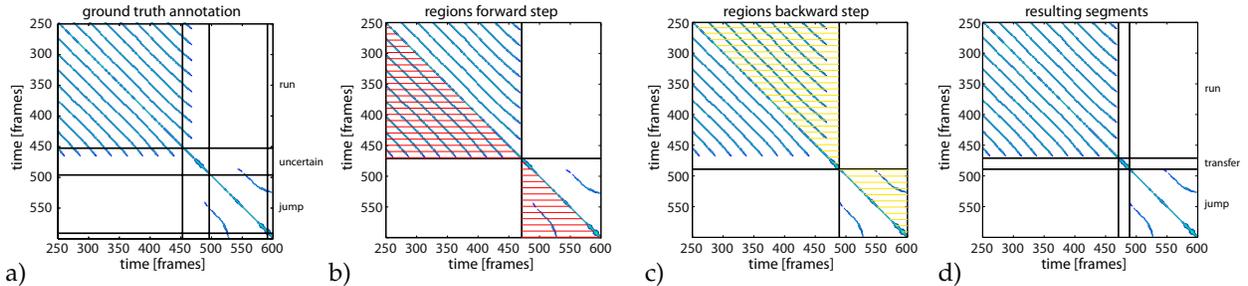
\begin{figure*}[t]
  \centering
  \includegraphics[width=0.9\linewidth]{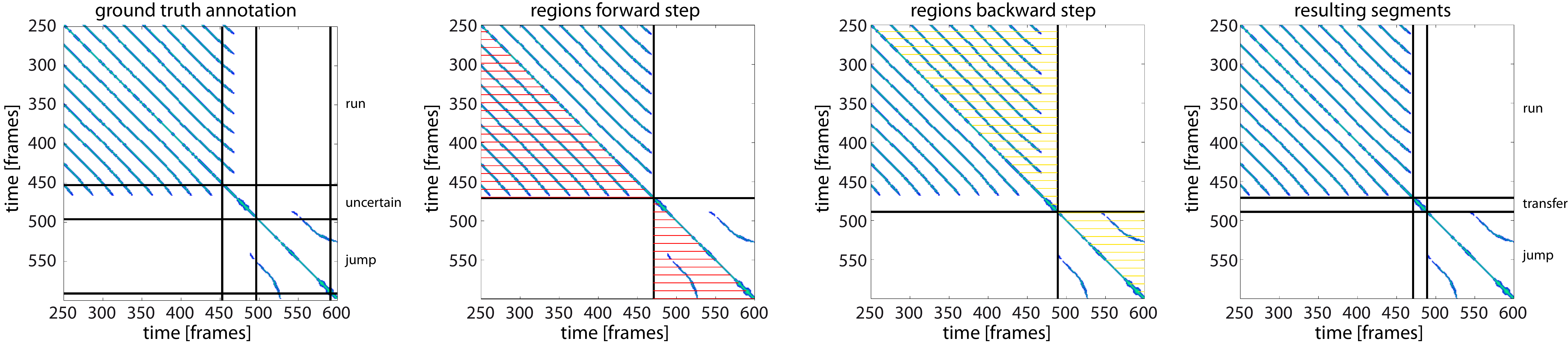}
  \setlength{\unitlength}{1cm}
    \begin{picture}(0,0)(8,0)
    	\put ( -8.6,-0.1){\mbox{a)}}
    	\put ( -4.3,-0.1){\mbox{b)}}
    	\put ( -0.1,-0.1){\mbox{c)}}
    	\put (  3.8,-0.1){\mbox{d)}}
    \end{picture}
  \caption{Region growing for activity separation: a) Sparse self similarity matrix with ground truth annotations; the label 'uncertain' indicates an area of inconclusive user annotations. b) Same matrix without main diagonal; also results for forward step of region growing. c) Results backward step of region growing. d) Final outcome of segmentation.
  Note that in the actual computation of the region growing steps (b) and c)) the main diagonal is removed. It is displayed in this representation for visualization purposes though.}
  \label{fig:sketch-selfsimmat}
\end{figure*}

The input to our method is a multi-dimensional time series acquired by recording a motion trial. 
First, the local neighbors of each frame in the motion sequence are found as a preliminary step. The sequence is then partitioned into distinct temporally coherent action segments; a subsequent step investigates the structure of these actions to find recurring patterns, i.e.\ shorter \textit{motion primitives} potentially enclosed as part of the action.

\subsection{Local Neighbors}\label{neighborhood}


A motion sequence $M$ is given as a collection of $m$ subsequent 
data points $p_1,\ldots,p_m$, each of which is represented by a feature vector $F=\left(f_1,\ldots,f_N\right)$ of dimension $N$ capturing the information over time. The features are modality specific and stacked in the time dimension, yielding a vector representation $[p_{i-f_1}, p_i, p_{i+f_2}]'$ of features over a window $w=[i-f_1,i,i+f_2]$ in a higher dimension.


Within the feature space, we define a search radius $r$ to search for the neighbors. Given no prior knowledge of the input data, we introduce a \textbf{generalized search radius} $R$ which is independent of the feature set time window size $w$ and input data dimensionality $N$.  $R$ is defined as the search radius for a window size of $|w|=1$ and dimensionality of $N=1$; the search radius is then rescaled as $r=R\sqrt{|w|\cdot N}$.

We construct a \textbf{kd-tree} from all feature points $F_i$ in the input stream and then search for the points located within the radius $r$ based on the Euclidean distance $d_{ij}$ between the features $F_i$ of $p_i$ and $F_j$ of $p_j$.  As a result of this search, we obtain a set $S_i$ of neighbors for each data point $p_i$.
The neighbors are specified as pairs $(j\in [1:M],d_{ij})$ of an index $j$ to a frame in the input motion and the distance $d_{ij}$ between the query point and neighbor $j$.

The sets of neighbors of a motion trial can be converted into a \textbf{sparse self similarity \replaced{matrix}{matrices}(SSSM)}.
\added{Self-similarity matrices are commonly used in human motion analysis for a range of tasks ranging from retrieval~\cite{KovarGleicher2004} to multi-view action recognition~\cite{Junejo2011}.}
A SSSM, as shown in Figure~\ref{fig:featurecondensation_sssm}~a), is generated by initializing an empty $M\times M$ matrix $\mathcal{M}$. For each frame $i\in [1:M]$ we set the entries $\mathcal{M}_{i,j} \forall k\in S_i$ to the values of $d_{ij}$ which are stored in $S_i$. An illustration of this connection is also given in Figure~\ref{fig:mat_graph_toy}~a) and~b).  Note that we use the matrices only for visualization in this work; for efficiency purposes, computations are performed directly on the sets of neighbors or derived data structures.

\subsection{Segmentation into Distinct Activities}\label{seg1}
Figure~\ref{fig:sketch-selfsimmat} shows an example SSSM with two cyclic activities, running and jumping, separated by an 'uncertain' period of inconclusive user annotations.  Note that the cyclic activities are characterized by structured diagonal blocks.  We separate the activities by searching for these characteristic blocks, using \textbf{region growing} to determine the blocks' borders.

A connected region starts as a seed in the upper left corner of the neighborhood representation matrix $\mathcal{M}_{1,1}$.  Contents of the lower triangular matrix below the main diagonal are then probed using scan lines. The triangular region is gradually extended to adjacent rows as long as the number of nearest neighbors in the updated region increases.
\replaced{If no new neighbors are found between frame $i$ and $i+w$ in the larger region, the current region is considered complete. The parameter $w$ is introduced to handle nosy data and is set to $w=8$ in all our experiments.}
{If no new neighbors are found at a frame $i$ in the larger region, the current region is considered complete.}
With such a stop criterion, neighbors from the main diagonal of the SSSM cannot be considered --- otherwise these elements would also be counted in the region-growing and result in a large region covering the entire SSSM.  We remove all entries of the main diagonal in the proximity corresponding to one second, based on observations that cyclic behavior in motion data does not occur at higher speeds.  A new region is then started from the upper left entry of the remaining matrix $\mathcal{M}_{i,i}$, with scan lines probing the content of the lower triangular matrix between $\mathcal{M}_{i,i}$, $\mathcal{M}_{i,j}$ and $\mathcal{M}_{j,j}$, where $j$ is the current frame.

The region growing is performed once as a \textbf{forward step}, seeding the first region at frame $\mathcal{M}_{1,1}$ of the matrix, to identify the end of repetitive patterns (see Fig.~\ref{fig:sketch-selfsimmat}b) and once as a \textbf{backward step}, seeding at the last frame $\mathcal{M}_{n,n}$ of the input sequence first to identify the \replaced{start of}{startof} the actions (Fig.~\ref{fig:sketch-selfsimmat}c). The lower right corners of the forward region growing correspond to end frames of an action, while the upper left corners of the backwards step correspond to the start frames of an action. Areas in between are considered transitions between the repetitive parts.

For efficiency, we work on the sets $S_i$ of neighbors, counting the number of entries in the neighborhoods between the seed frame and the current scan line index.  Because we work with a symmetric matrix, this is equivalent to scanning triangular parts of the sparse similarity matrix.

Compared to the region growing approach of V\"ogele et al.~\cite{voegele2014a}, where the neighbors were counted in a quadratic region, the method presented here is more computationally efficient. For each scan line, only one set of nearest neighbours needs to be considered, while for the  quadratic region in~\cite{voegele2014a}, all preceding neighbour sets are reconsidered for each growing step. The runtime complexity is therefore reduced from $O\left(k\frac{n(n-1)}{2}\right)$ to $O(k·n)$ in our approach, where $k$ is the maximum number of nearest neighbors and $n$ is the number of frames of the motion trial---in the worst case the first region grows over the whole SSSM.

%

\subsection{Subdividing Actions into Motion Primitives}\label{seg2}
Once the input sequence is segmented into distinct actions, we can search for motion primitives within each action. Consider a single action, such as walking in Figure~\ref{fig:sketch-repetitions}; we want to find the reoccurring units, i.e.\ steps of which the activity consists. Such units are responsible for the minor diagonals in the SSSM of the specific activity, with start and end frames of each unit corresponding to the start and end position of a diagonal.  
Rather than searching for the diagonals' starts and ends in the SSSM, we use an alternative neighborhood graph representation and simplify the problem to finding the shortest path. \added{We note that if no reoccuring primitives are found, the action segment is considered itself a single primitive.}

\subsubsection*{Alternative Representation by a Neighborhood Graph}\label{sec:LNGconstruction}
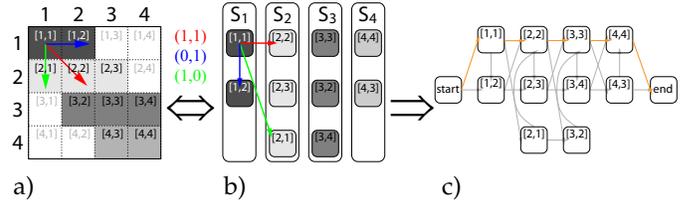
\begin{figure}[t]
	\centering
	\includegraphics[width=1\linewidth]{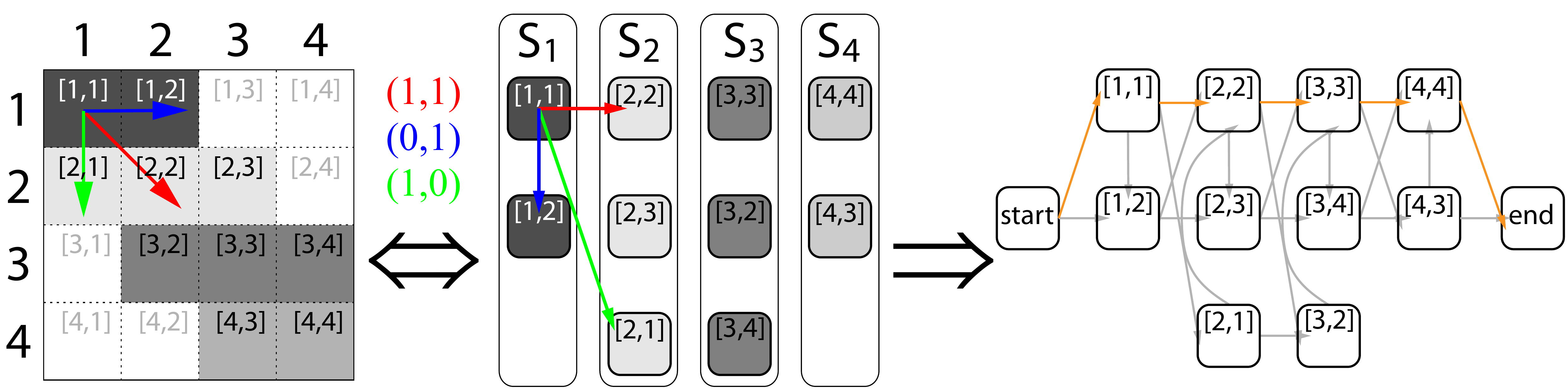}\\
	\setlength{\unitlength}{1cm}
	\begin{picture}(0,0)(4,0)
	\put ( -0.4 ,0){\mbox{a)}}
	\put (  2.4 ,0){\mbox{b)}}
	\put (  5.3 ,0){\mbox{c)}}
	\end{picture}
	\caption{Toy example illustrating the relationship between SSSM and neighborhood graph $G_M$. a): SSSM of four consecutive points with allowed steps indicated by arrows. Red arrow: step $(1,1)$, blue arrow: step $(0,1)$, green arrow: step $(1,0)$. b): Neighborhoods of each of the four points, e. g. $S_i$ is the neighborhood of the point $i$. c): Resulting neighborhood graph.}
	\label{fig:mat_graph_toy}
\end{figure}

All of the neighbors $p_j$ within one specific activity stored in the set of neighborhoods $S=\{S_i,\ i \in 1,\ldots,n\}$ can be considered as nodes of a graph $G_\text{act}$. The criteria for connecting the nodes in this graph is based on accessibility between the points and can be characterized by the concept of dynamic time warping.
Consider two points $p_j \in S_i$ and $p_{j^{\ast}}\in S_{i^{\ast}}$. A valid time warping step to access the point $p_{j^\ast}$ from $p_j$ is defined as a pair  $(a,b) \in \{(1,1),(0,1),(1,0)\}$ such that $p_{j^\ast} =p_{j+b}$ and $S_{i^{\ast}} = S_{i+a}$. In particular, the point $p_{j^\ast}$ is always an entry further below in the neighborhood list $S_{i^\ast}$ than $p_j$ is in the list $S_i$, while $S_{i^\ast}$ is either identical to $S_i$ or lies to its right hand side ($S_{i^\ast} = S_{i+1}$).
Figure \ref{fig:mat_graph_toy} shows a toy example to illustrate a possible scenario.

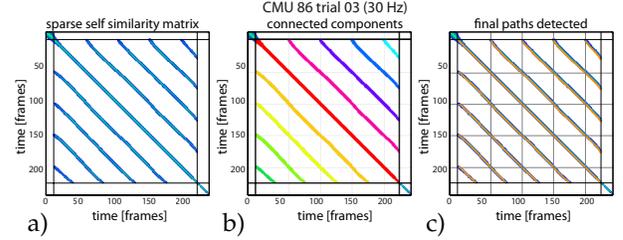
\begin{figure}[t]
	\centering
	\includegraphics[width=0.9\linewidth]{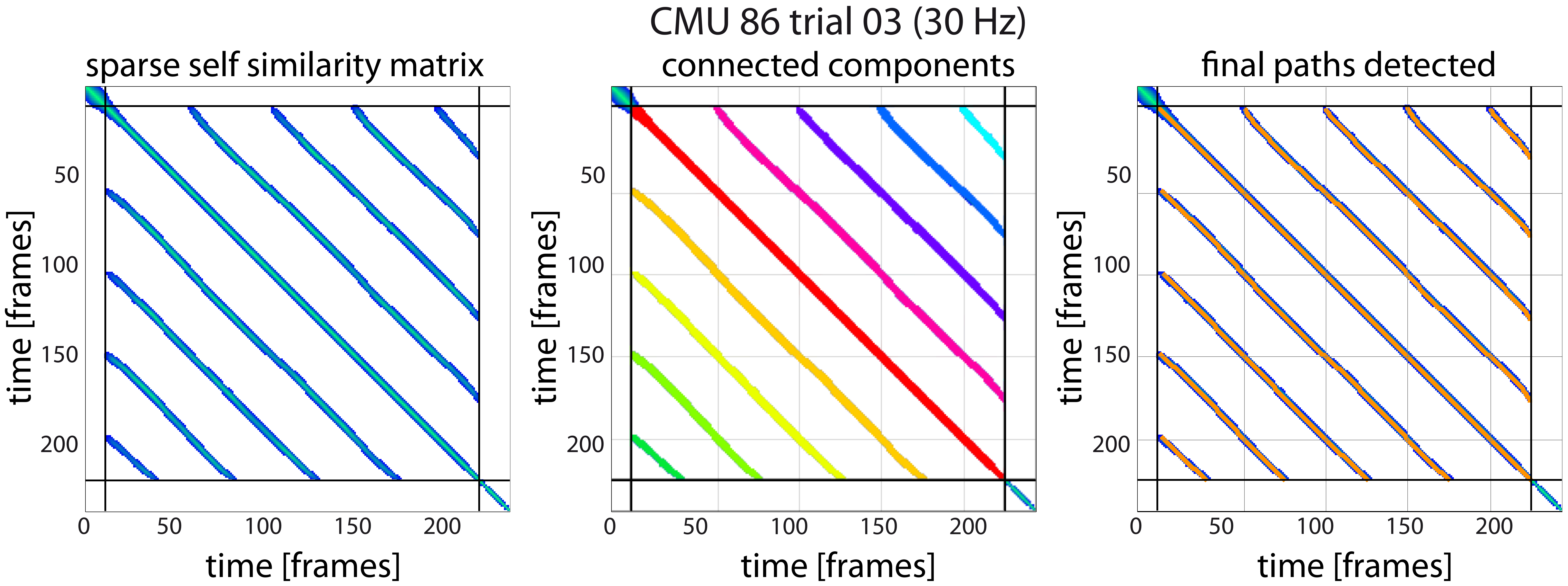}
	\setlength{\unitlength}{1cm}
	\begin{picture}(0,0)(8,0)
	\put ( 0.0,-0.1){\mbox{a)}}
	\put ( 2.6,-0.1){\mbox{b)}}
	\put ( 5.3,-0.1){\mbox{c)}}
	\end{picture}
	\caption{Illustration of connected components in $G_{\text{act}}$. a) SSSM corresponding to walking. b) Same matrix with its connected components color coded. c) Optimal warping paths highlighted by orange lines.}
	\label{fig:sketch-repetitions}
\end{figure}

An important property of the graph $G_\text{act}$ which we exploit lies in the following observation: each diagonal of the matrix $\mathcal{M}_{\text{act}}$ reflecting local similarities is one \textbf{connected component} of $G_\text{act}$. For the next steps, it is useful to work only with neighbors belonging to the same connected component $cc$ for a given frame, with the resulting restricted graph denoted as $G_{cc}$ (Refer to Figure~\ref{fig:sketch-repetitions} for a visualization of connected components).

\subsubsection*{Computation of Warping Paths}\label{sec:LNGpathsearch}

Dynamic time warping, to calculate an optimal match between two given time series $A$ and $B$ with certain restrictions, creates a path between these sequences. The sequences are matched non-linearly in the time dimension to optimize for a similarity measure. Technically, a warping path $\mathcal{P}_{A,B}$ of length $\lambda$ between two such sequences is given as a pair of vectors $(v_A,v_B)$ where $v_A = (a_1,\ldots, a_\lambda)$ with $a_i\in A$ meeting constraints such as $a_i\leq \nu a_{i+1}$ for all indices and $v_B = (b_1,\ldots,b_\lambda)$ with $b_i\in B$ meeting $b_i\leq \nu b_{i+1}$. In our experiments, we use $\nu = 2$. The constraints on $v_A$ and $v_B$ could be seen as upper and lower limit of the paths \textbf{slope}.

The sets of neighbors ${S_i}$ are suitable to replace conventional dynamic time warping based on the neighborhood graph described above.
(for more details on how this can be used instead of time-warping, see Kr\"uger et al.~\cite{Krueger10} using the example of motion capture time series).

Since each diagonal in the SSSM translates to one connected component in the graph, searching for an optimal warping path reduces to finding a shortest path through the connected component $G_{cc}$. To this end, we add one additional start and one end node to the graph. The start node connects to all nodes corresponding to the first set of neighbors in the component, while the end node is connected to all nodes that correspond to the last set of neighbors. Now, the warping path can be found efficiently by searching the shortest path from the start to the end node. The costs of a path is the accumulated distance of the included neighbors.

We further limit the set of warping paths per activity based on their length and slope. First, paths covering less than five frames of the motion trial are discarded; such paths are found for very short but similar segments existing between longer primitives.  Although these segments may be semantically meaningful, we ignore these to prevent extremely short primitives.  Secondly, paths with average gradients less than 0.5 and larger than 2 are also discarded. Such paths represent mapping between motions whose speeds vary by a factor greater than two. We want to avoid such cases, e.g.\ when a longer standing sequence is mapped to a few poses in the middle of a walking cycle.

For each valid warping path $\mathcal{P}_i$ we have a pair $(a_i,b_i)$ representing the starting position within the SSSM. These positions correspond to the bordering frames between motion primitives.
We only check if any candidates are closer than 5 frames. If this is the case we only consider the one where the corresponding warping path had smaller costs.

\subsubsection*{Complexity Analysis}
The critical computation step for detecting primitives is building the graph representation from the sets of nearest neighbors. Creating this graph requires checking all possible connections of each neighborhood entry in $\mathcal N$ to other entries by a number of $s$ possible steps. For each of the neighborhoods of each activity there is a maximum of $k$ entries. Since the number of edges is limited by $O(k·s·n)$, the search for connected components is limited to the same complexity, the  overall run time complexity is also $O(k·s·n)$.


\section{Symmetry of Motion Data}\label{motion_symmetry}




Motion data contains several intrinsic symmetries which can be exploited during analysis. We focus on mirrored motions and begin by defining the plane of symmetry.  Let $X=\left\{x_1,\ldots, x_J\right\} \in \mathbb R^{3\times J}$ be a geometric model of a moving subject, i.e. an ordered set of joints. 
A motion of $X$ is a multi-dimensional trajectory $\mathbb X$ of $X$ over time. Let $P_X$
be a plane spanned by two perpendicular vectors which connect joints or linear combinations of joints.

For human models, the plane of symmetry is the saggital plane, i. e. the vertical plane which passes from anterior to posterior, dividing the body into left and right.
A motion is symmetric with respect to this mirror plane if, for a set of descriptive features $\mathcal F$, at least one pair of features $f_i, f_j \in \mathcal F$ can be switched without imposing a (significant) change on $\mathbb X$.
The concept of the mirror plane can be transferred from humans to other models; all vertebrates are bilaterally symmetrical with two pairs of appendages such as limbs, fins or wings and the saggital plane is also a mirror plane.  



The symmetry of an action segment $X_A$, based on its mirrored version $X'_A$ mirrored at the saggital plane, may be characterized as follows:
\begin{enumerate}
	\item\label{it:motsym1} $X_A$ is highly symmetric if its primitive segmentation is exactly the same as that of $X'_A$.
	\item\label{it:motsym2} $X_A$ is exclusively phase-shift symmetric if its primitive segmentation has no cuts in common with $X'_A$.
	\item\label{it:motsym3} $X_A$ is asymmetric if the primitive segmentation of $X'_A$ returns no cuts at all.
\end{enumerate}




Naturally, a mixture of two or more situations is possible when an action contains different types of motion primitives (see Appendix~\ref{app:symmetry} for an overview). Therefore, it makes sense to treat each square region of the SSSM (identified by the activity separation) representing individual motion activities separately, as there may be some activities which have symmetric counterparts and some which do not.  
%
%
We make use of phase-shifted symmetry in order to distinguish phase-shifted primitives like single steps in walking.


\section{Clustering of Motion Primitives}\label{sec_clustering}
By clustering the motion primitives, it is possible to find the frequency with which the same primitives occur and also have an indication of the semantic and temporal relationships between different primitives. This is of great interest for motion synthesis, using motion graphs~\cite{MinChai2012}, or for motion analysis, in terms of action recognition. We propose a unique clustering algorithm in which we do not have to provide the number of clusters in advance.

A cluster graph $\mathbf{G}_{\mathbb M}$ is used to store the similarity information between the motion primitives. In this cluster graph, each primitive is represented as a node.
Consider now a sparse self similarity matrix associated with a motion $M$; the motion primitives $m_q$ are represented by squares on the main diagonal. The goal is now to search for valid warping paths between each pair of motion primitives $m_i$ and $m_j$.
To this end, we can build a neighborhood graph (see Section~\ref{sec:LNGconstruction}) including the neighbors in the rectangle region that is spanned when comparing $m_i$ and \replaced{$m_j$}{$\mathcal{m}_j$}.
The shortest path from the top entries to the bottom entries in this submatrix is found if it exists. If this shortest path satisfies a minimum length requirement and has a slope inside the range of valid slopes (see Section~\ref{sec:LNGpathsearch}) we add an edge between the corresponding nodes in $\mathbf{G}_{\mathbb M}$.
After the algorithm has gathered all similarity information, a search for the strongly connected components is performed on $\mathbf{G}_{\mathbb M}$.
Each strongly connected component represents a cluster of motion primitives.

The algorithm presented above is a modified version of the algorithm of V\"ogele et al. \cite{voegele2014a}.
Both approaches perform equally well for clustering, though the current approach is more efficient, since only small neighborhood graphs between each pair of primitives are constructed, while in the previous work one large neighborhood graph was constructed over all neighbors of the trial and then was cut into smaller parts when needed.

\section{Experiments}\label{sec:experiments}


We now report on a series on experiments to show the effectiveness of our approach.
First, we compare our results with previous methods on a set of motion capture data.
Second, we show that meaningful results are obtained when using different sensor modalities such as accelerometers and EMG sensors.
Finally, we apply our approach to Kinect skeleton data and show that our motion primitives are meaningful and consistent with of human-annotated key frames.

\begin{figure*}[t]
  \centering
  \includegraphics[width=0.95\textwidth]{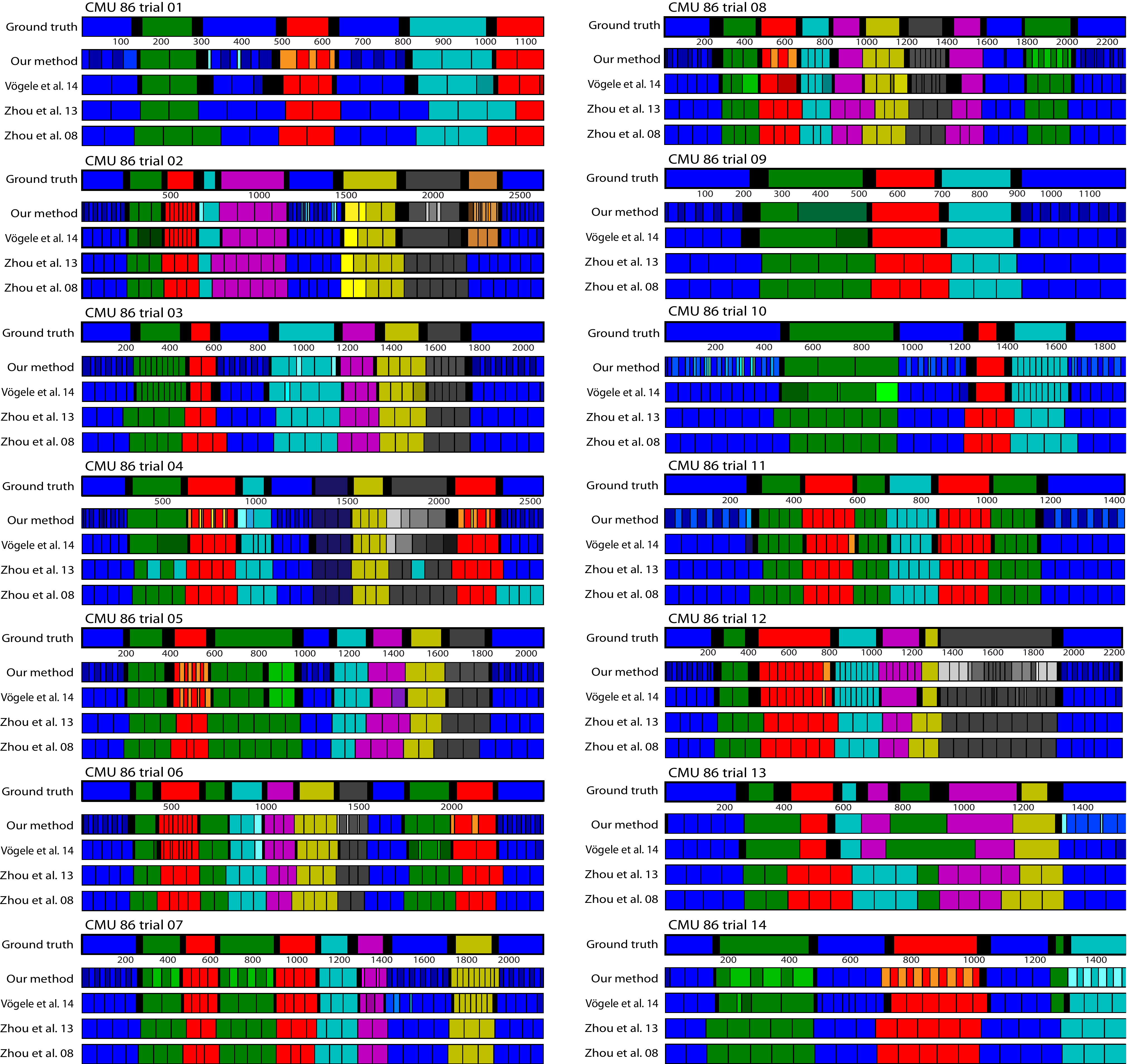}
  \caption{Segmentation results for CMU 86 trial 01 to 14. For each of the 14 trials, the first row displays the human ground truth annotations, the second row compares them to our results. The results of the (H)ACA methods (Zhou et al.~\cite{Zhou08,Zhou13}) and V\"ogele et al.~\cite{voegele2014a} are given in the two lower rows. Note that there is a variety of units of different sizes indicating that the actual length of motion primitives may vary considerably.}
  \label{fig:res_86_01-14}
\end{figure*}

\subsection{Segmenting Markered Motion Capture Data}\label{exp:mocap}
We apply our segmentation method to the the motion sequences of subject 86 of the CMU database~\cite{mocapcmu}, as was done by\added{ }
Zhou et al~\cite{Zhou08,Zhou13}.
We compare our segmentations to theirs, as well as to our previous work~\cite{voegele2014a} in Figure~\ref{fig:res_86_01-14}.
We show a number of improvements in comparison to~\cite{Barbic04, Zhou13}, the most notable being the ability to segment fine-structured motion primitives in nearly all cases.  First, we are able to distinguish different styles of executing a task.  For example, in the case of wiping a window/black board (CMU, trial $12$ of subject $86$), Zhou et al.~\cite{Zhou13} group all primitives together as the same type (dark grey blocks), while we are able to distinguish between back and forth wiping motions versus circular wiping movements (various shades of grey).

Secondly, we are able to distinguish symmetric movements and separate primitives accordingly.  Examples include rotation of the body in trial $7$ (variations of blue and green blocks), dribbling the ball (right vs. left hand) in trial $14$ (dark green, light green and red, orange blocks).  Other \replaced{approaches}{approach} are unable to distinguish between a step forward with the left versus right foot, nor handling of the ball with the left versus right hand.



\begin{figure}[t]
  \centering
    \setlength{\unitlength}{1cm}
    \begin{picture}(0,0)(0,0)
    	\put (0.1,0){\mbox{b)}}
    	\put (0.1,4.3){\mbox{a)}}
    \end{picture}
  \includegraphics[width=0.45\textwidth]{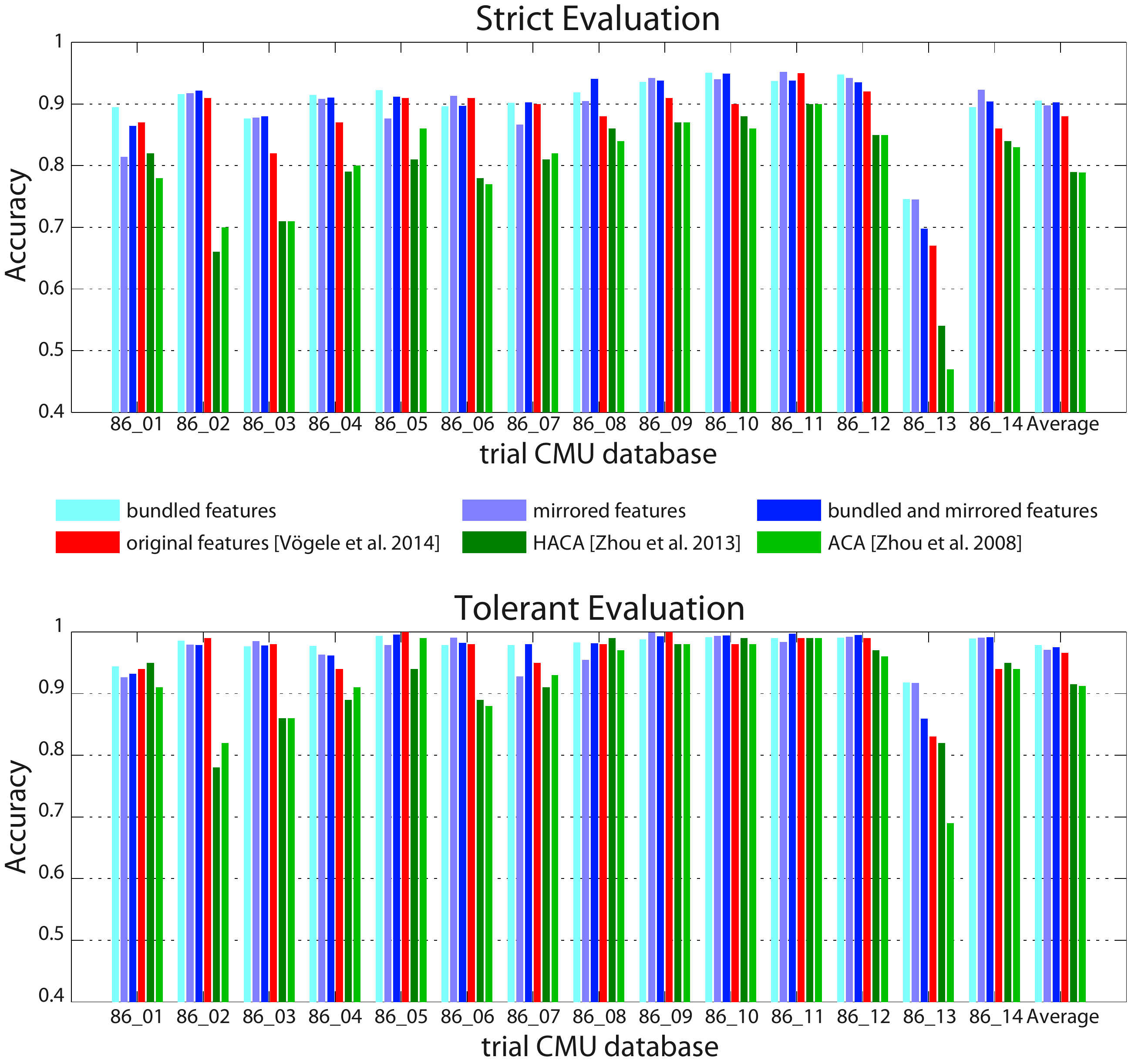}
  \caption{Accuracy values of different segmentation methods: Light blue bars refer to our method (bundled features only), purple: our method (mirrored features only), blue: (bundled and mirrored features combined), red: V\"ogele et al.~\cite{voegele2014a}, dark green: Zhou et al.(ACA)~\cite{Zhou08} light green: Zhou et al.(HACA)~\cite{Zhou13} using (a) the strict evaluation, counting strictly the classes all methods detect and (b) the more tolerant evaluation which allows transitions as valid classes.
  For both evaluations the bundled and mirrored features give higher accuracy values in average compared to the original features and the (H)ACA based approaches. Especially the bundled features have a better effect on the accuracy compared to the mirroring.}
  \label{fig:eval_strict}
\end{figure}
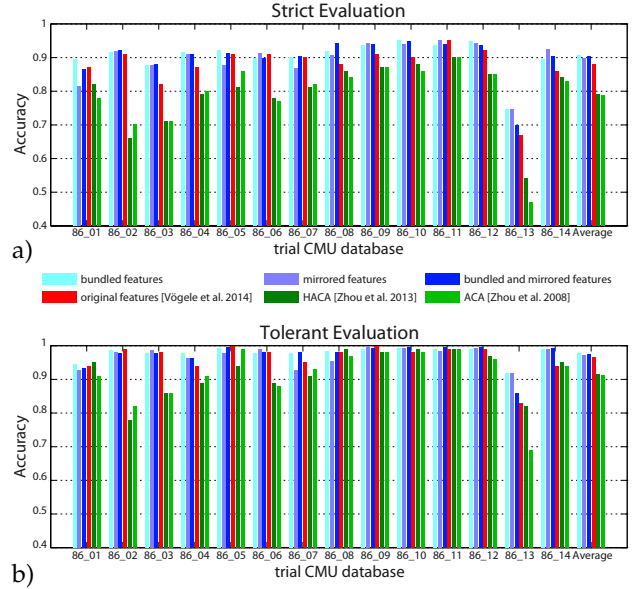

\begin{figure}[t]
  \centering
    \setlength{\unitlength}{1cm}
    \begin{picture}(0,0)(0,0)
    	\put (0.1,0){\mbox{a)}}
    	\put (5.3,0){\mbox{b)}}
    \end{picture}
  \includegraphics[width=0.45\textwidth]{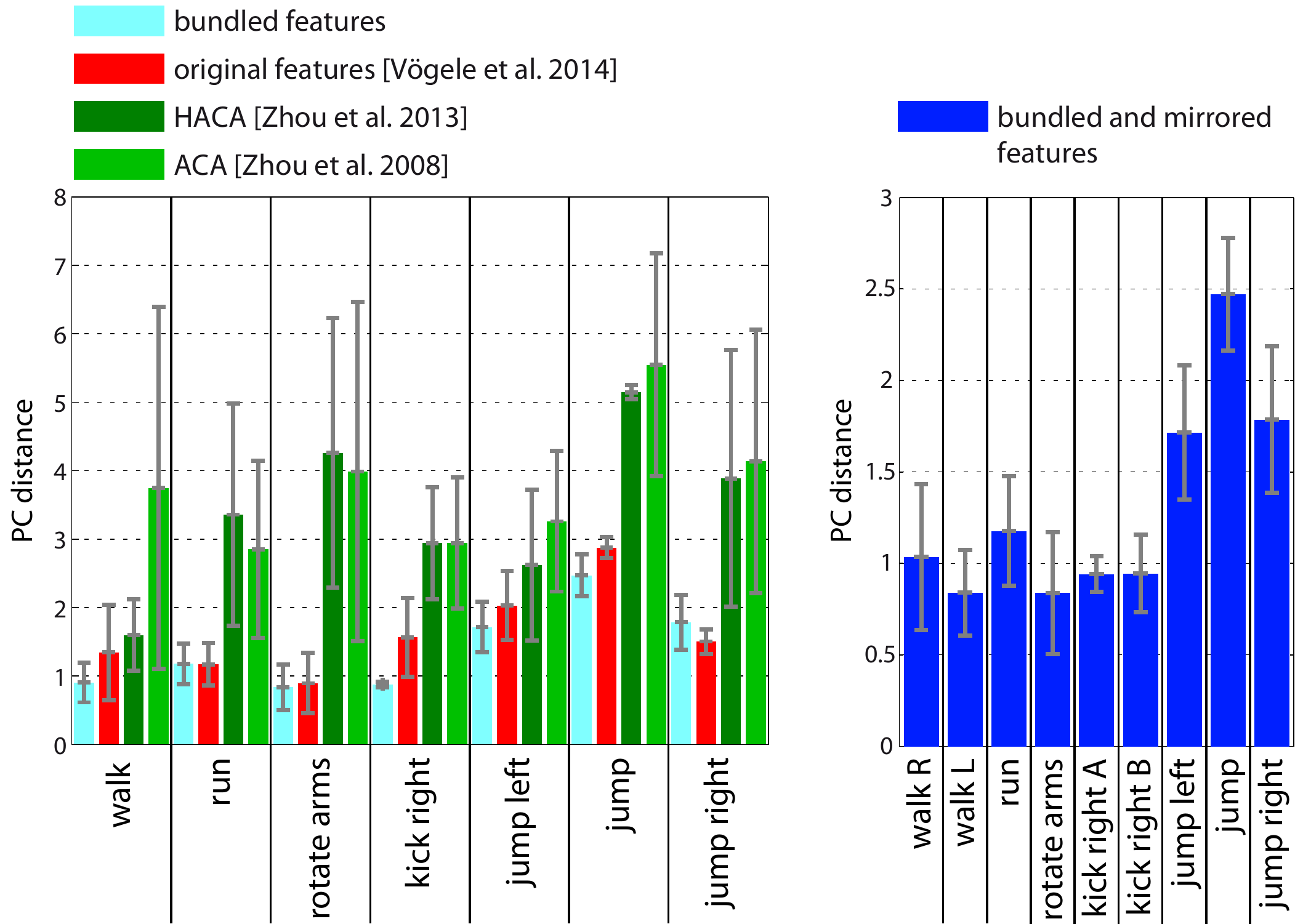}
  \caption{a) Evaluation of clusters produced by our method, the method of V\"ogele et al., the HACA and the ACA method. The example at hand is trial $3$ of CMU subject $86$. The respective means are given by the blue color bars for our method, red for V\"ogele et al., (dark) light green color bars for (H)ACA, with variance shown as error bars. Note that lower distances reflect more consistent clusters.
  b) Evaluation of clustered results produced by our method. 
  Here we plot the mean D, with variances shown as error bars as in (a) but for the case where both bundling and symmetry features are included. This is one case where additional motion classes are introduced by exploiting symmetry of motion: there are two classes of steps and also two classes of 'kicks'. The distinctions are caused by different types of symmetry: walking is phase-shifted and kicking included one part which was symmetrical (more static) and one which was asynchronous.
  }
  \label{fig:clusterVarComp}
\end{figure}


\subsubsection*{Accuracy Comparison}
Our method produces the same action classes as 
\cite{voegele2014a} and nearly the same classes as \cite{Zhou08,Zhou13}.  We use the same methods as \cite{voegele2014a} to evaluate the segmentation accuracy on a frame-wise level, using both a strict and a more tolerant evaluation, and present the results in Figure~\ref{fig:eval_strict}. For a motion primitive $s$, the strict method checks whether all of $s$'s frames belong to the same action class as the other primitives found from the same segment; the tolerant method eases the constraint to both the same action class as well as transition/uncertainty segments.
Due to the finer division of primitives found by our method, there are different clusters representing the same motion. For example, walking consist of alternating left and right steps. For consistent evaluation with previous methods, we have assigned such symmetrical counterparts to the same class, i.e.\ the 'left step' cluster and the 'right step' cluster are both assigned to the walk action.  We achieve significantly higher accuracy values for both types of evaluation, with an average of $90\%$ for our method, $88\%$ for the method of V\"ogele et al., and $79\%$ for (H)ACA for the strict evaluation and $99\%$ in comparison to $97\%$ for V\"ogele et al., $92\%$ for HACA and $91\%$ for ACA for the tolerant evaluation. \added{V\"ogele et al. discuss the application of the segmentation technique to the label transfer problem. Based on these values we anticipate that applying feature bundling would yield similar results.}

\subsubsection*{Intra-Cluster Variance for Motion Primitives}

Dynamic time warping is an established distance measure for temporal sequences and accumulates the local (frame-wise) distances from one segment to the warped version of the other. For a given cluster $C$ a cumulative distance measure $D$, tallied over all pairs of segments $s_i$ and $s_j$ contained in the cluster can be defined as:
\begin{equation}
\small{D = \sum_{i=1, j\neq i}^{\|C\|} \left(\frac{\mathrm{DTW_\alpha}(s_i,s_j)}{\|s_i\|}\right)}
\end{equation}
where $\|s_i\|$ is the length of segment $s_i$ and $\mathrm{DTW_\alpha}$ is the DTW distance of point clouds as defined by Kovar et al.~\cite{KovarGleicher2002}. Note that $D$ is particularly sensitive to outliers and will detect scattered or inconsistent clusters.

By making finer distinctions between motion primitives, we achieve lower intra-cluster variances for the clusters.  
Figure~\ref{fig:clusterVarComp}~a) compares the clustered results by our method, the method of V\"ogele et al., the HACA~\cite{Zhou13} and the ACA~\cite{Zhou08} method. 
Figure~\ref{fig:clusterVarComp}~b) shows the same variance values for the case where mirrored features are included. The classes show similarly low variance. The given example is one case where additional motion classes are introduced by exploiting symmetry of motion: there are two classes of `steps' and also two classes of `kicks'. The distinctions are caused by different types of symmetry. While walking is phase-shifted, kicking included one part which was symmetrical (more static) and one which was asynchronous (where the leg was up).

Our clusters group together a variety of motion primitives without transitions and primitives from other actions.  In particular, our primitives reflect exactly the number of repetitions within actions. For illustration, there are five repetitions of 'rotate arms' in one sequence (see Figure~\ref{fig:res_86_01-14}, Subject 86 trial 03, frames $1600$-$1800$) and we segment this into exactly five primitives. Zhou et al.'s methods~\cite{Zhou08,Zhou13} does not account for the inherent repetition and segments the action into three primitives, thereby yielding much higher DTW distances between these primitives.

Over all clusters of all trials from actor 86 we obtain a average cluster variance of 1.18 (min: 0.52, max: 2.66, std: 0.42) for mirrored and bundled features, 1.17 (min: 0.58, max: 2.58, std: 0.37) for bundled features, 1.69 (min: 0.69, max: 3.45, std: 0.54) for the original features. Compared to HACA 2.51 (min: 0.48, max: 8.78, std: 1.53) and ACA: 2.56 (min: 0.86, max: 9.42, std: 1.50).

\subsubsection*{Parameter Evaluation}
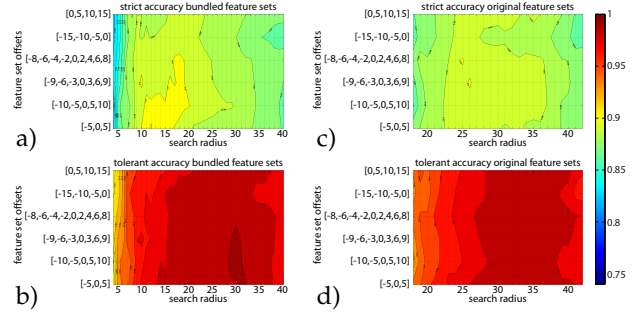
\begin{figure}[tbp]
\centering
\setlength{\unitlength}{1cm}
\includegraphics[width=0.45\textwidth]{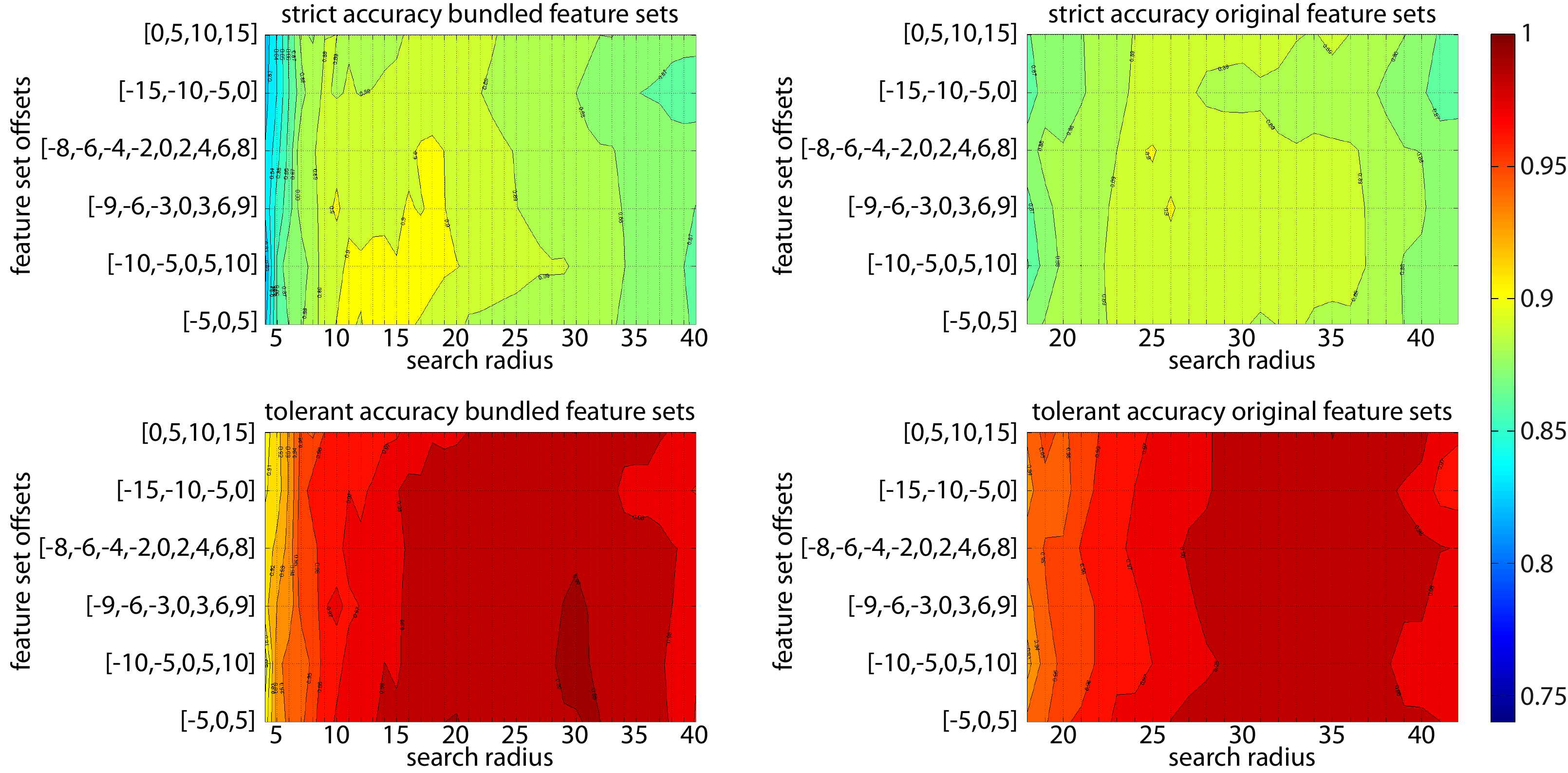}
\begin{picture}(0,0)(0,0)
    \put (-8.2 ,2.1){\mbox{a)}}
    \put (-8.2 ,0.0){\mbox{b)}}
    \put (-4.2 ,2.1){\mbox{c)}}
    \put (-4.2 ,0.0){\mbox{d)}}
\end{picture}
\caption{Maps displaying the accuracy values for various combinations of the parameters: The generalized radius $R$ is plotted on the horizontal axis, choices of stacking offsets are plotted along the vertical axis. A stacking offset of $[-5, 0, 5]$ is a concatenation of frames at times $i-5$, $i$ and $i+5$.  Parts a) and b) shows the maps based on the bundled features, where a) is the stricter and b) is the more tolerant version. Parts c) and d) show the same based on the bundled features, where c) is the stricter and d) is the more tolerant version.}
\label{fig:parameters}
\end{figure}

The most important parameters for our segmentation method are the search radius (Section~\ref{seg1}) and the temporal window for feature stacking (Section~\ref{neighborhood}). We found that the method is insensitive to either parameter and show the segmentation accuracy in Figure~\ref{fig:parameters} for various parameter settings for  the strict and tolerant evaluations for both original and bundled features.
All plots show that the accuracies are high for nearly all possible combinations of parameters.

\begin{figure}[t]
\setlength{\unitlength}{1cm}
\includegraphics[width=0.48\textwidth]{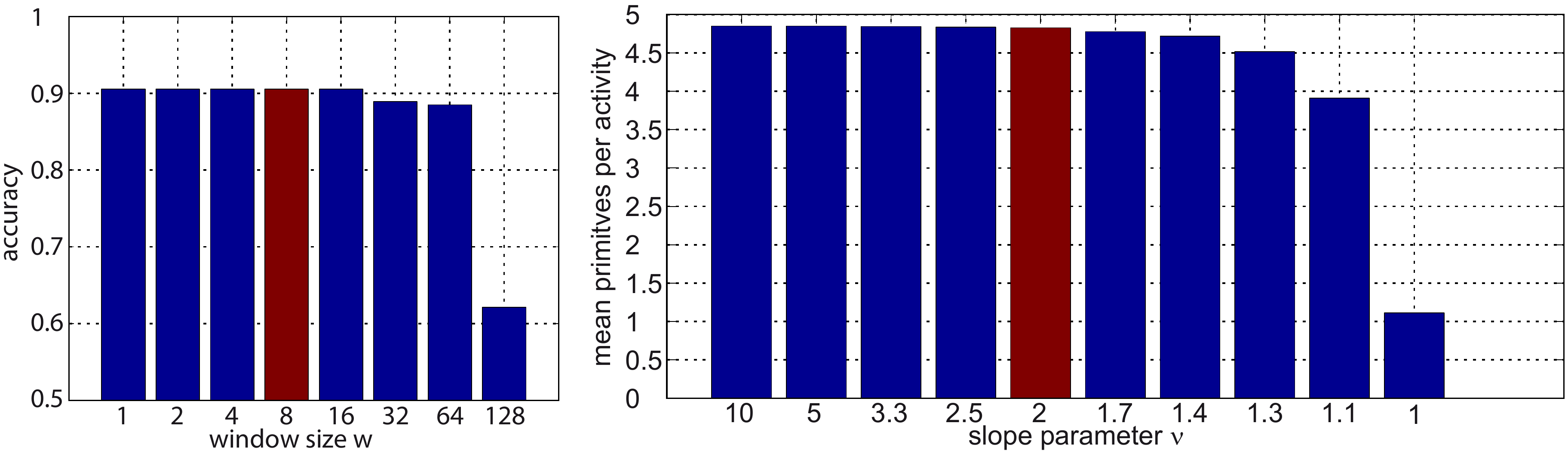}
\begin{picture}(0,0)(0,0)
    \put (-8.8 ,-0.1){\mbox{a)}}
    \put (-5.5 ,-0.1){\mbox{b)}}
\end{picture}

\caption{
\added{
a) Accuracy (strict evaluation) for varying window sizes $w$ of the region growing step. $w=8$ (red bar) was chosen for our experiments.
b) Average number of motion primitives per activity based on the minimal and  maximal slope of the warping paths. $\nu=2$ (red bar) was chosen for all other experiments in this work.}
}
\label{fig:parameter_slope}
\end{figure}
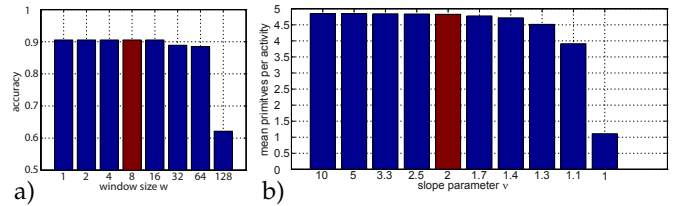

\bk{
Our region growing in the activity separation step stops if no new neighbours were found in a window of $w$ frames.
We tested our approach with various window sizes and computed the strict accuracy measure for evaluation. Figure~\ref{fig:parameter_slope}~a) shows the accuracy results. The method is very robust against this parameter, the accuracy only drops when the window size is chosen extremely large (128 frames).

For segmenting the motion primitives, there are the additional parameters of the allowed minimal and maximal value of the warping path slope $\nu$ (Section~\ref{sec:LNGpathsearch}). Following conventions in the literature~\cite{KovarGleicher2004, Krueger10} we set the slope to be within the window $\frac{1}{\nu}$ and $\nu$.
In all our experiments we set this parameter to be $\nu=2$. We evaluated this parameter by computing the average number of motion primitives found per activity on the motion capture dataset. The results are shown in Figure~\ref{fig:parameter_slope}~b). If $\nu$ is smaller than 2 the number of primitives drastically decreases, while larger values for $k$ don't increase the number of motion primitives. We choose the value of two to permit extreme temporal deformations between motion segments. Although, our experiments indicate that such warps did not occur in the CMU dataset.
}

\subsubsection*{Details on Timings}

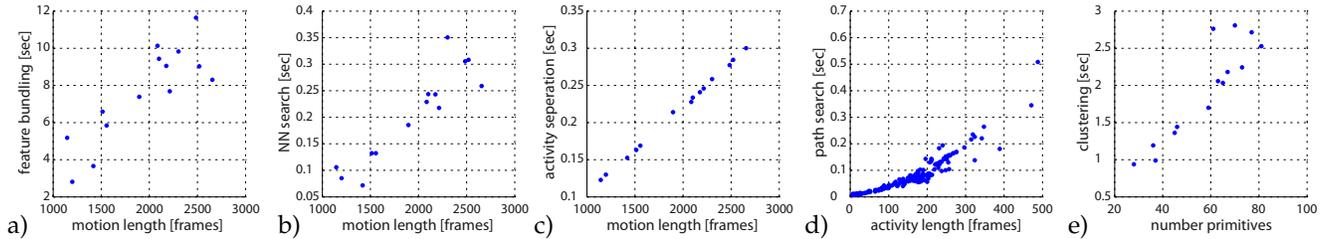
\begin{figure*}[t]
  \centering
  \includegraphics[width=0.95\textwidth]{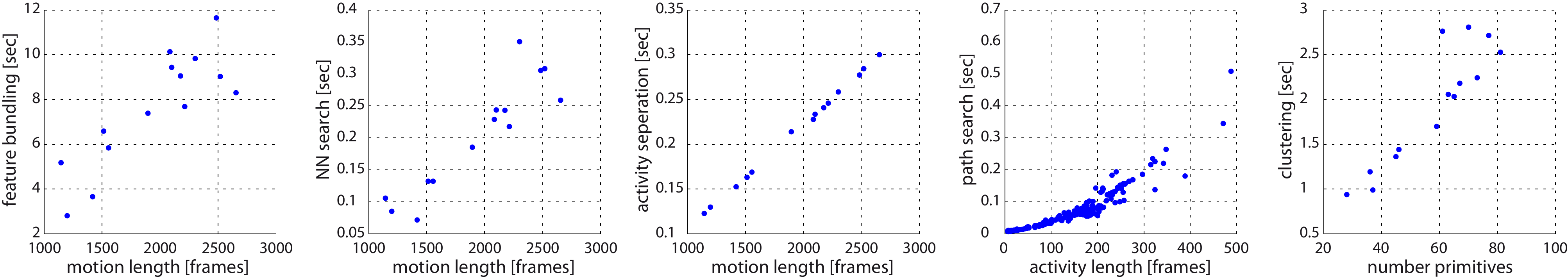}
	  \setlength{\unitlength}{1cm}
    \begin{picture}(0,0)(4,0)
    	\put ( -13.6 ,0){\mbox{a)}}
    	\put ( -10 ,0){\mbox{b)}}
    	\put ( -6.6 ,0){\mbox{c)}}
        \put ( -3 ,0){\mbox{d)}}
    	\put ( .5 ,0){\mbox{e)}}
    \end{picture}
  \caption{Scatter Plots for timings of the steps of our method. a) feature bundling, b) knn search, c) region growing (scan lines), d) path searches, e) clustering. Note that the complexity of the first three steps (a-c) is linear in the number of frames, whereas path searches (d) are quadratic in the lengths of the activities. In practice, the clustering (e) also computes linear in the number of motion primitives.}
  \label{fig:timings_steps}
\end{figure*}

Figure~\ref{fig:timings_steps} shows a timing breakdown for the CMU examples. The first segmentation step (part c)) including feature bundling (part a) and knn search (part b) activity detection are, in practice, approximately linear in the number of frames (Figure~\ref{fig:timings_steps} a) - c)). In theory, the worst case complexity for region growing is $O(kn)$, i.e.~ when the first region grows from the first to the last frame of the input motion sequence.  This case was not observed in practice.
The shortest path searches for primitive detection (part d)) is quadratic with respect to the number of frames per activity. Finally, the clustering step (part e) also computes in linear time (in the number of motion primitives).

On an Intel Core i7 4930K at 3.40GHz we were able to segment and cluster each motion sequence (up to 3000 frames in length) in less than $15$ seconds using our single threaded Matlab implementation.

\subsection{Combined Motion Sensors}
To demonstrate our algorithm's effectiveness on different sensor modalities, we recorded a set of  electromyography (EMG) and acceleration motion data using a Delsys Trigno wireless acquisition system. EMG recordings show electrical activity produced by skeletal muscles and are commonly analyzed in biomechanics and neurology.
Acceleration recordings show the local accelerations due to changes of the sensors velocity and are commonly analyzed in biomechanics and sport sciences. Nearly all `wearables' use accelerometer readings to analyse user activity.
Typically, analysis of EMG and accelerometer signals is done on sequences already segmented into motion cycles; the segmentation is almost always done manually, and frequently relies on other readings such as motion capture or video data. Our automatic segmentation technique can significantly improve the work flow in domains using EMG and accelerometer recordings.

\begin{figure}[t]
  \centering
  \includegraphics[width=0.45\textwidth]{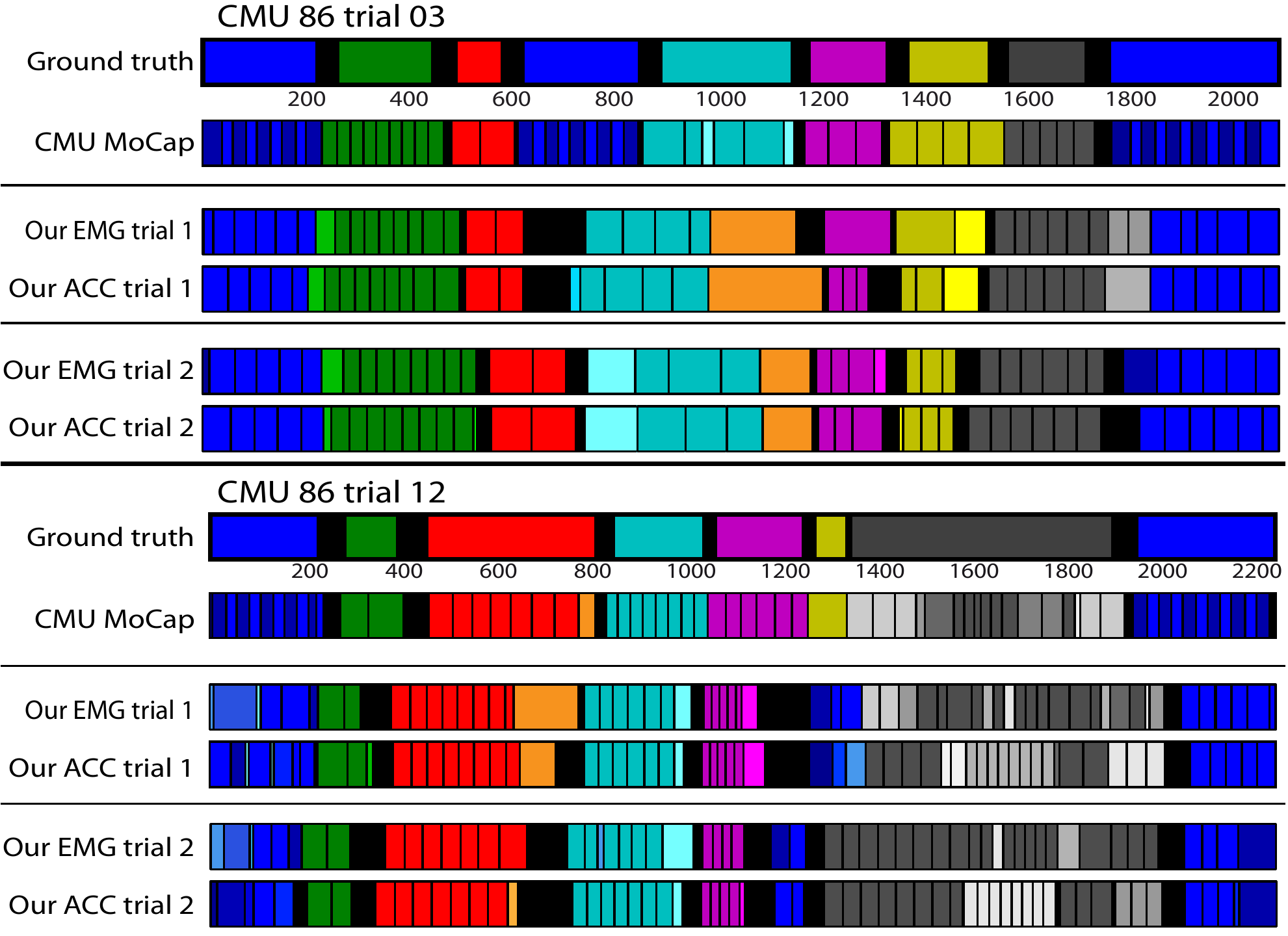}
  \caption{Exemplary results of two trial As can be seen from the key point visualization, this third example is also an interesting situation for the consistency measure.s of subject $86$ of the CMU. The color codes correspond to the different motion clusters the primitives were assigned to by our method. The results are very similar between the data sets, even between the original and the re-enacted recordings.}
  \label{fig:emg_acc_results}
\end{figure}

We take recordings from one trial subject who was asked to re-enact the sequences of subject $86$ in the CMU database. Results of two repetitions of trials $3$ and $12$ are  compared to the CMU originals (see Figure \ref{fig:emg_acc_results}). The raw EMG readings consist of data streams of $16$ sensors, each documenting the activation of large skeletal muscle groups including the larger flexors and extensors of the human body (refer to Appendix~\ref{datasources} for documentation).

All EMG recordings were pre-processed in a standard fashion: the signals were rectified, re-sampled from $2000$ Hz to $30$ Hz and smoothed by a $20$ Hz low pass filter (see \cite{Chowdhury2013} for a review on EMG data processing). Acceleration recordings were re-sampled from $120$ Hz to $30$ Hz as well and filtered by a binomial filter over a window of $16$ frames.

The segements and primitives shown in Figure~\ref{fig:emg_acc_results} show that our approach works on EMG and acceleration data as successfully as clean motion capture data, despite the former two being much noisier input sources.
The EMG segments are very similar to the acceleration segments representing the same motion; in most actions, the same number of primitives were found across the two modalities.
We hypothesize that the slight differences in timing are due to inherent differences in the signals being captured by the two modalities.

Clustering of the motion primitives is comparable between the two modalities for most activities. The largest differences are in `wiping a window' (grey blocks in trial 12). Here we were unable to distinguish between back and forth wiping versus circular wiping in the EMG, while the accelerations gave clear motion primitive clusters. The EMG data does not reflect this difference since the muscle activation on the main arm extensors and flexors do not change as clearly as the accelerometer readings.

\subsection{Kinect Action Data}\label{kinectAction}
Processing noisier markerless motion capture systems can be \added{a} challenge, but
our method is able to deal with such data in a reliable way when the feature bundling step is included. We segment the MSRC-12 Kinect Gesture Data Set from Fothergill et al.~\cite{msrc12}, consisting of a number of action sequences which were originally recorded for action recognition. The data set consists of $594$ sequences in total from $17$ actors. The trials range from $1000$-$2000$ frames recorded at $30$ frames per second. The data are available as 
$35$ joint angles of $3$ scalars over a length of $n$ frames.
Examples of different segmentation results can be found in Figure~\ref{fig:examples_Kinect}.
\subsubsection*{Impact of Feature Bundling}

We show the original input data streams, as well as our segmentation results, color-coded according to our clustering, both with and without feature bundling. The given key points~\cite{Nowozin2012} annotate the actual gestures at a specified key frame and are indicated by red lines in all subplots. Feature \replaced{bundling allows}{bundlingallows} us to segment a regular pattern of primitives that coincide with the time series (see Figure~\ref{fig:examples_Kinect}(a) and (b)).
Figure~\ref{fig:examples_Kinect}(b) shows some more variation in the lengths of the primitives. This originates from a break in the input motion (first half of the trial, darkblue bar) and also some speed variation (later half of the trial) which can clearly be seen in the plot of the data stream. Figure~\ref{fig:examples_Kinect}(c) is particularly interesting because the primitive have a much finer structure. The motion primitives occur as ones recurrent groups of smaller parts (light green/green) alternating with a longer primitive (blue) and are well aligned with the keypoint annotations.

Without feature bundling, the segmented motion primitives are far less regular. In Figure~\ref{fig:examples_Kinect}(a), three repetitions were broken up into two individual parts (yellow and turquoise). In Figure~\ref{fig:examples_Kinect}(b), the longer breaks (brown) between the repetitions were found but not all repetitions (dark blue) could be cleanly separated.  Finally, in Figure~\ref{fig:examples_Kinect}(c), a similar pattern was found both with and without feature bundling, but without bundling, the shorter primitives are split into smaller irregular parts that are assigned to different clusters.

\subsubsection*{Consistency Evaluation}
\begin{figure}[t]
  \centering
   \setlength{\unitlength}{1cm}
      \begin{picture}(0,0)(0,0)
    	\put (0.2,3.5){\mbox{a)}}
    	\put (4.4,3.5){\mbox{b)}}
        \put (0.2,0){\mbox{c)}}
    	\put (4.4,0){\mbox{d)}}
    \end{picture}
  \includegraphics[width=0.45\textwidth]{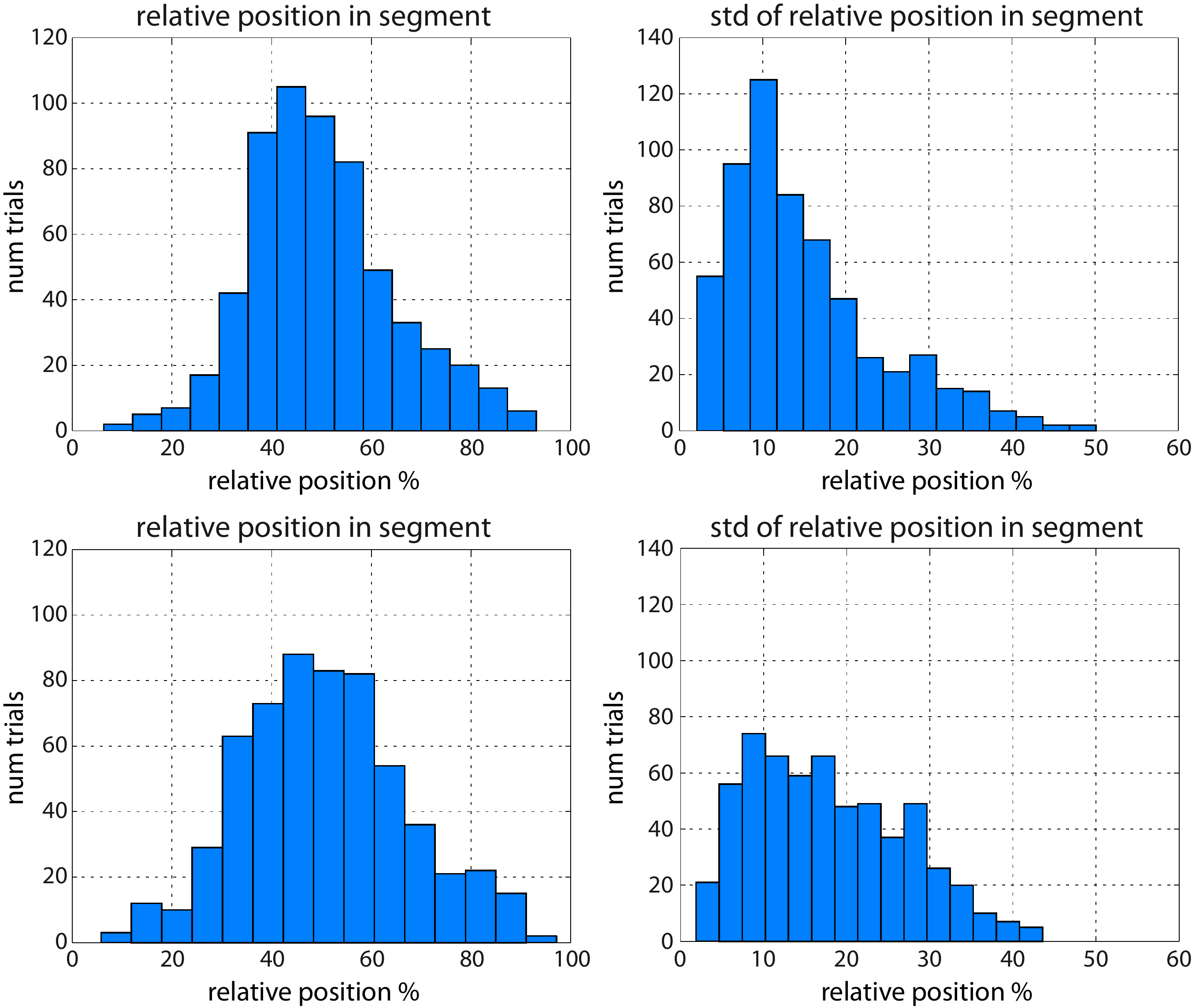}
  \caption{The histograms show the average location of given key point annotations (left hand side) in the motion primitive and their standard deviations (right hand side). In the majority of cases, the location of key points is approximately at the center of the corresponding primitive with a deviation of less than $20\%$ if feature bunding is used a) and b). If the original features are given as input the average locations are spread with larger standard deviations c) and d).}
  \label{fig:kinect_eval}
\end{figure}
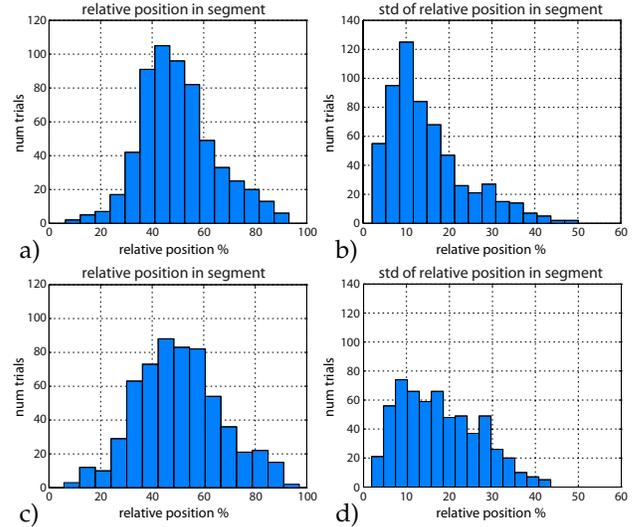

We measure how well our identified motion primitives correspond to the given keypoint annotations.
To this end, for each of the motion primitives, 
we measured the difference between its start frame and the key point and scale this value proportionally to the length of the primitive and plot the distribution as well as the standard deviations in Figure~\ref{fig:kinect_eval}.
The histograms shows that the key-point annotation as judged by the human annotator occurs consistently at a relative position of 40-60\% in the segmented motion primitive (see Figure~\ref{fig:kinect_eval}(a)), with a relatively low standard deviation of $15\%$ around the mean values (see Figure~\ref{fig:kinect_eval}(b)) when feature bundling is enabled.
Figure~\ref{fig:kinect_eval}(c) and (d) show the same values when the segmentation is done without feature bundling; here the mean values are more spread with higher standard deviations.

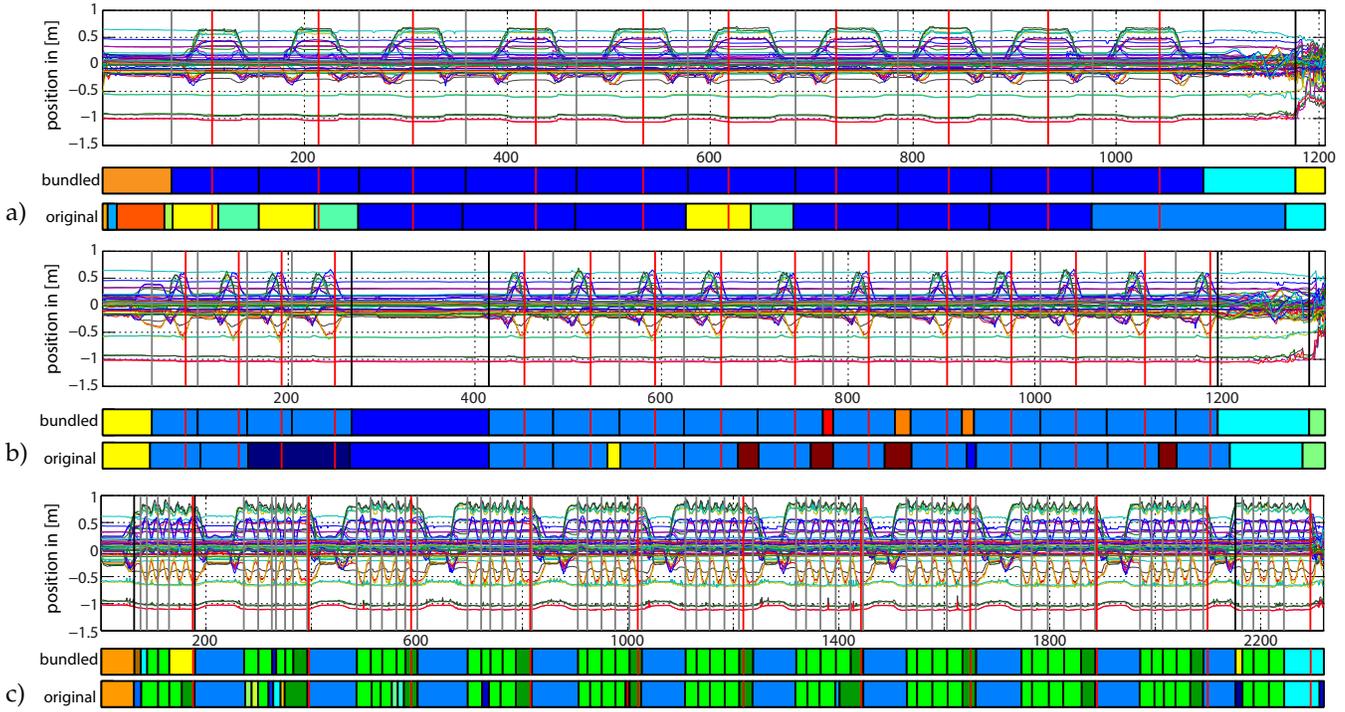
\begin{figure*}[t!]
\centering
\setlength{\unitlength}{1cm}
\includegraphics[width=0.95\textwidth]{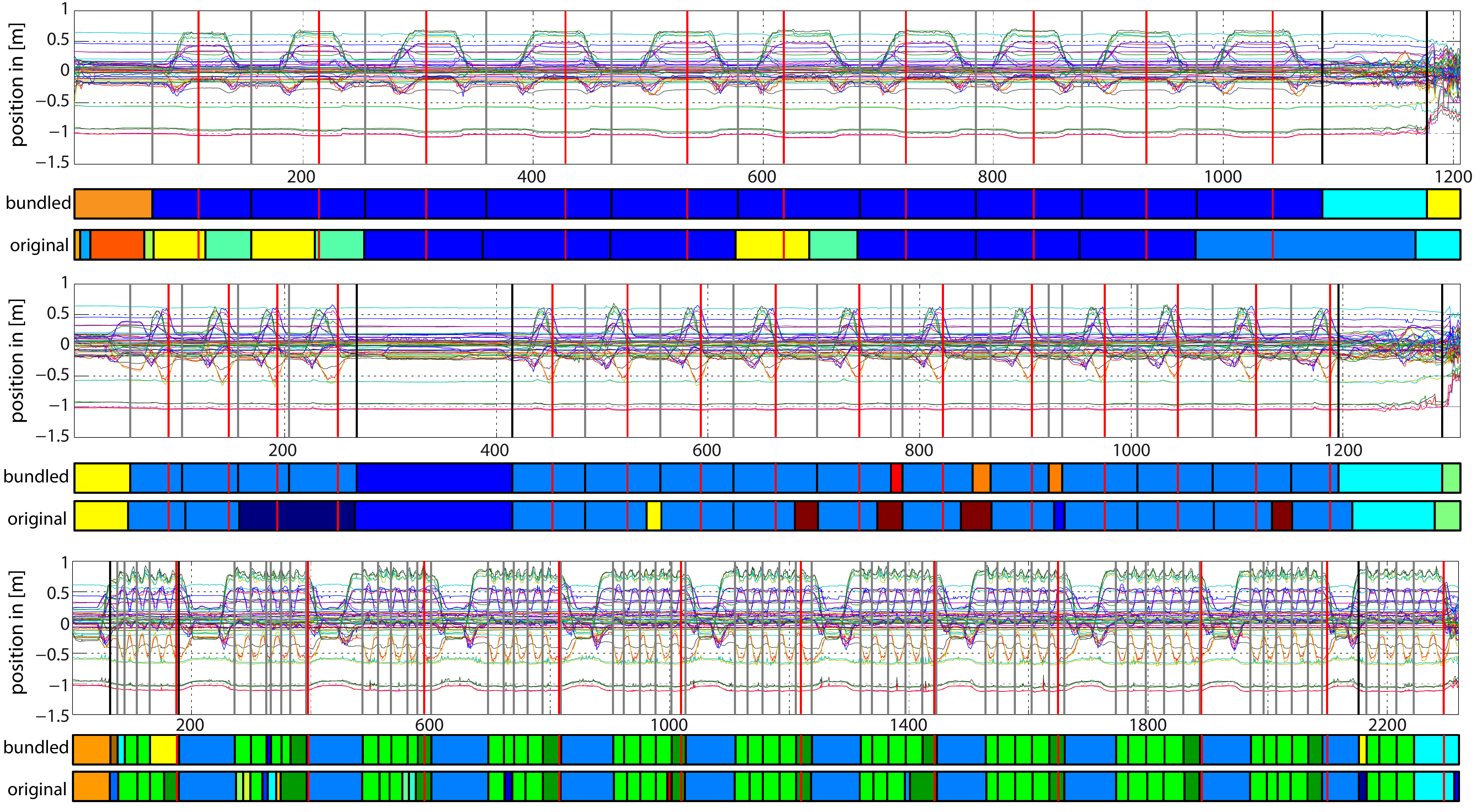}
\begin{picture}(0,0)(0,0)
    \put (-17.8 ,6.6){\mbox{a)}}
    \put (-17.8 ,3.4){\mbox{b)}}
    \put (-17.8 ,0.2){\mbox{c)}}
\end{picture}
\caption{Three different results produced on Kinect recordings. The input time series are plotted for an overview of the general structures of the underlying motions, showing motions with (a) regular primitives segments [P1\_2\_g09\_s08] (b) a sequence with a longer break and also some speed variation [P1\_2\_g05\_s08] and (c) many quick repetitions of primitives, i. e. waving both hands (see green bars)[P3\_2\_g11\_s29]. The very fine primitive segmentation is desirable for motion understanding, but may be punished by the consistency measure based on key point locations since the key point annotation now falls at the end or beginning of the fine primitives.}
\label{fig:examples_Kinect}
\end{figure*}

While the results are very consistent with respect to the given annotations when the bundled features are used, there are still a number of exceptions which contribute to the distribution of deviations. One example is shown in in Figure~\ref{fig:examples_Kinect} (c)) where a sequence of smaller motion primitives occurs arranged in the same order.  According to our goal, to identify repetitive motion primitives, the fine segmentation is the desired result. However, this is non-ideal for the consistency measure, since it considers the relative position within the motion primitive. With very fine primitives, the key points are at the end of the last small cluster (green) or at the beginning of the subsequent larger cluster (blue).  Grouping together smaller primitives, in conjunction with re-clustering could alleviate this problem.

\subsection{Non-cyclic Kinect data}
\begin{figure*}[t!]
\centering
\setlength{\unitlength}{1cm}
\includegraphics[width=0.95\textwidth]{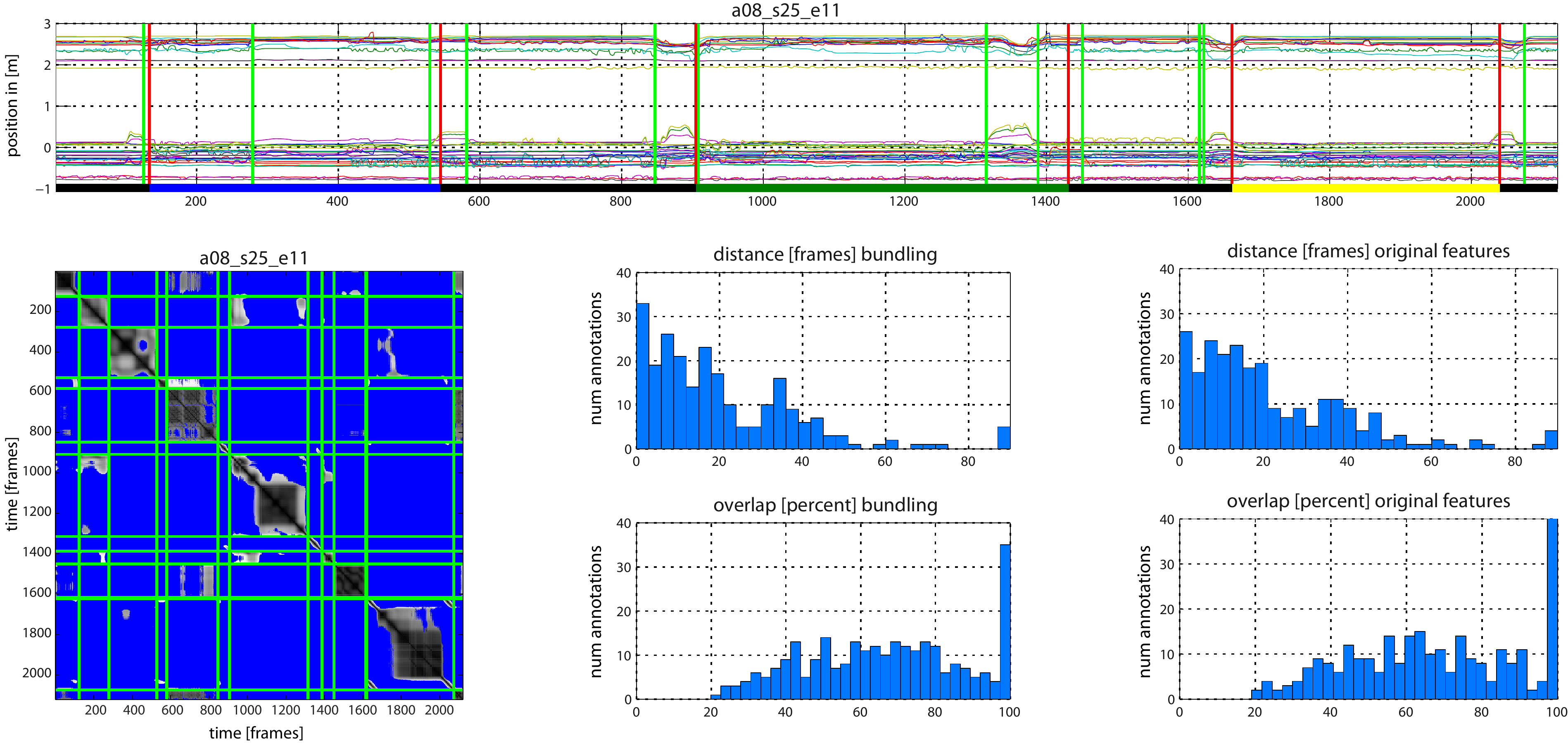}
\begin{picture}(0,0)(0,0)
    \put (-17.8 ,6.1){\mbox{a)}}
    \put (-17.8 ,0.0){\mbox{b)}}
    \put (-11.3 ,3.0){\mbox{c)}}
    \put (-5.3  ,3.0){\mbox{d)}}
    \put (-11.3 ,0.0){\mbox{e)}}
    \put (-5.3  ,0.0){\mbox{f)}}
\end{picture}
\caption{\added{Results produced on Kinect recordings from the MSR 3D Action Dataset. Segmentation results (green lines) and ground truth annotations (red lines) for trial a08\_s25\_e11 (a). The bars at the bottom indicate the ground truth labels: no action (black), drinking (blue), eating (green), reading book (yellow). The corresponding SSSM with segmentation results as green lines (b). Histograms for the distance between annotations and computed cuts on bundled (c) and original features (d). Histograms showing the overlap between annotated segments and primitives identified by our method on bundled (e) and original features (f).}}
\label{fig:MSRAction}
\end{figure*}
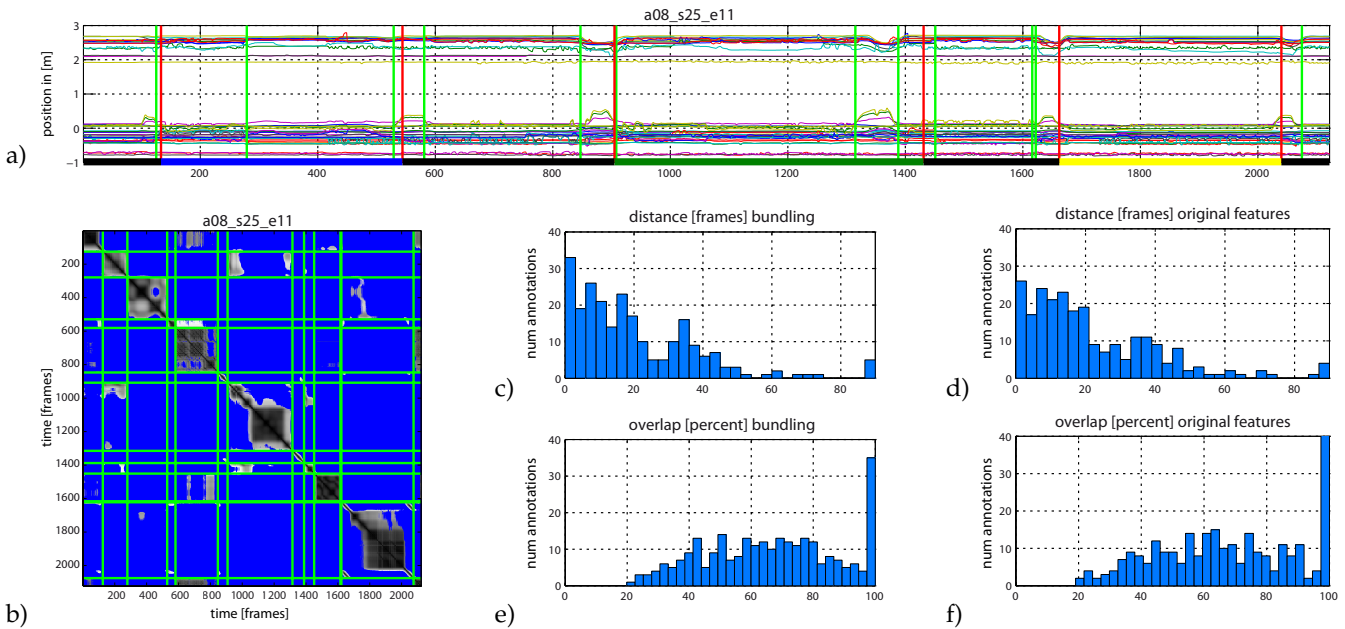

\added{
We tested our approach on the more challenging  MSR 3D Online Action Dataset~\cite{Yu2015}. This data set contains noisy, mostly non-cyclic actions such as drinking, eating, using laptop, reading cellphone,
etc. recorded with a Kinect device. We apply our method on the subset S4, which contains 36 sequences of long trials (1000-3500 frames recorded at 30 fps) from 11 subjects and is intended for continuous action recognition.  The data are available as 20 3D joint positions.
An example segmentation result can be found in Figure~\ref{fig:MSRAction}~(a) along with the corresponding SSSM~(b).
Since the annotations are given as intervals in this dataset we use the following measures to evaluate the quality of our results. First we measure the absolute distance in frames from the start and end points of the annotation to the next found cut. This measure alone would favor over-segmentation; thus we compensate with an additional measure of the overlap of an annotation with the largest segment found by our method .
Using bundled features we obtain a mean distance between start and end points of the annotations and our primitives
of 21.2 frames with a standard deviation of 19.6 frames.  The mean overlap is 66.56 percent with a standard deviation of 21.16. If we use original features we obtain 22.5 (mean) 19.7 (std) and 67.2 (mean) 22.3 (std), respectively. Figure~\ref{fig:MSRAction}~(c)-(f) shows the histograms for these values.
Since these sequences contain relatively static poses, the effect of feature bundling is not as strong as in the more dynamic data sets,  since there are less variations in static poses.
We were able to determine that many of the annotated actions in this data set can be split into three segments: a dynamic part in the beginning (e.g. raising cup to mouth), a static part in the middle (drinking) and another dynamic part in the end (lowering cup).
This corresponds to the rough annotations in this data set where complex actions are annotated as a block, while our approach returns finer motion segments.
}

\section{Conclusions}\label{sec:limitations}

We have presented a segmentation method which is able to process a number of data modalities and separate cyclic activities and their transitions. Our approach tries to tackle the segmentation problem on a general level in terms of the choice of crucial parameters, e.g. the search radius and the feature offsets for stacking. The feature bundling is a novel contribution is this area and has proven to be especially helpful for processing noisy data modalities \bk{such as EMG, accelerometer and Kinect motion capture. We used a five-point derivation to estimate the direction of movement in the bundling, but when faced with severe noise, one will need more robust methods. This will further reduce variance in the feature space, but there are few implications as long as one does not try to synthesize new sequences from the feature space.

So far, we have only shown our segmentation on sequences taken in fairly constrained settings.  We anticipate that it is also applicable to sequences taken "in the wild", given the right features and similarity measures.  This remains an open topic for future work.}  Challenges also remain to be seen once the segmentation is even further generalized, for example to spatial segmentation problems such as mesh animation sequences~\cite{sattler-2005-compression}.

Since our method is based on self-similarities, the limits \added{of finding primitives} are reached when input sequences contain only non repetitive activities (e.g. one step, jump, turn).  However, the assumption that most human activities are of repetitive nature is valid for motion capture data within existing data sets such as CMU~\cite{mocapcmu}\deleted{and HDM05}.
\deleted{Futhermore, we believe that combining our approach with an efficient lookup of previously segmented motion trials should make it feasible to segment even non-repetitive actions.}
For other sensor modalities we found a similar behaviour; if the motion is cyclic, so are the measured local accelerations.
For the EMG signals we could not identify any substantial changes between the individual repetitions in the muscle activation patterns.
Here changes may occur from fatigue effects in longer motion trials though this was not observed in our experimentation.

Currently, all tested data modalities have been processed individually; a natural next step is to jointly process several data types in a multi-modal setting. Combining the findings on more than one recording of the same motion could help refine segmentation and the resulting analysis.  Secondly, exploring smooth embedding approaches~\cite{Yao2011} for feature bundling is a promising direction for future work. Even though a PCA works for our examples, small artifacts are nonetheless visible, which may not be the case in other non-linear (but more computationally expensive) embeddings.

We offer source code for our feature bundling, segmentation and clustering approach online\footnote{\url{http://cg.cs.uni-bonn.de/en/projects/gemmquad/motionsegmentation/}}.  We believe that it will be of use for many different research applications and hope that it will encourage others in the community to work on generalized segmentation.

\bibliographystyle{IEEEtran}
\bibliography{2015_TPAMI_Segmentation}

%

\begin{IEEEbiography}[{\includegraphics[width=1in,height=1.25in,clip,keepaspectratio]{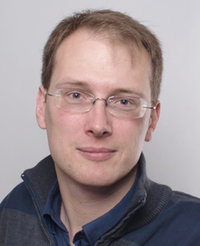}}]{Bj\"orn Kr\"uger}
studied computer science, mathematics and physics at Bonn university. He received his MS in computer science (Dipl.-Inform.) in 2006 and his PhD (Dr. rer. nat.) in computer science in 2012. From 2012 to 2015 he worked as postdoc at Bonn university. Since 2015 he joined the Gokhale Method Institute (Stanford, CA) as senior researcher.
His research interests include: computer animation, computer graphics, machine learning, and motion capture.
\end{IEEEbiography}

\begin{IEEEbiography}[{\includegraphics[width=1in,height=1.25in,clip,keepaspectratio]{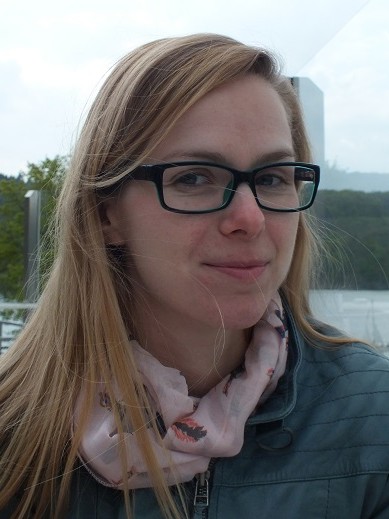}}]{Anna V\"ogele}
is a research assistent with the department of computer science at the University of Bonn. She received her MS in Mathematics (Dipl.-Math.) from the University of Bonn in 2011. Her research interests include computer graphics and animation as well as machine learning.
\end{IEEEbiography}

\begin{IEEEbiography}[{\includegraphics[width=1in,height=1.25in,clip,keepaspectratio]{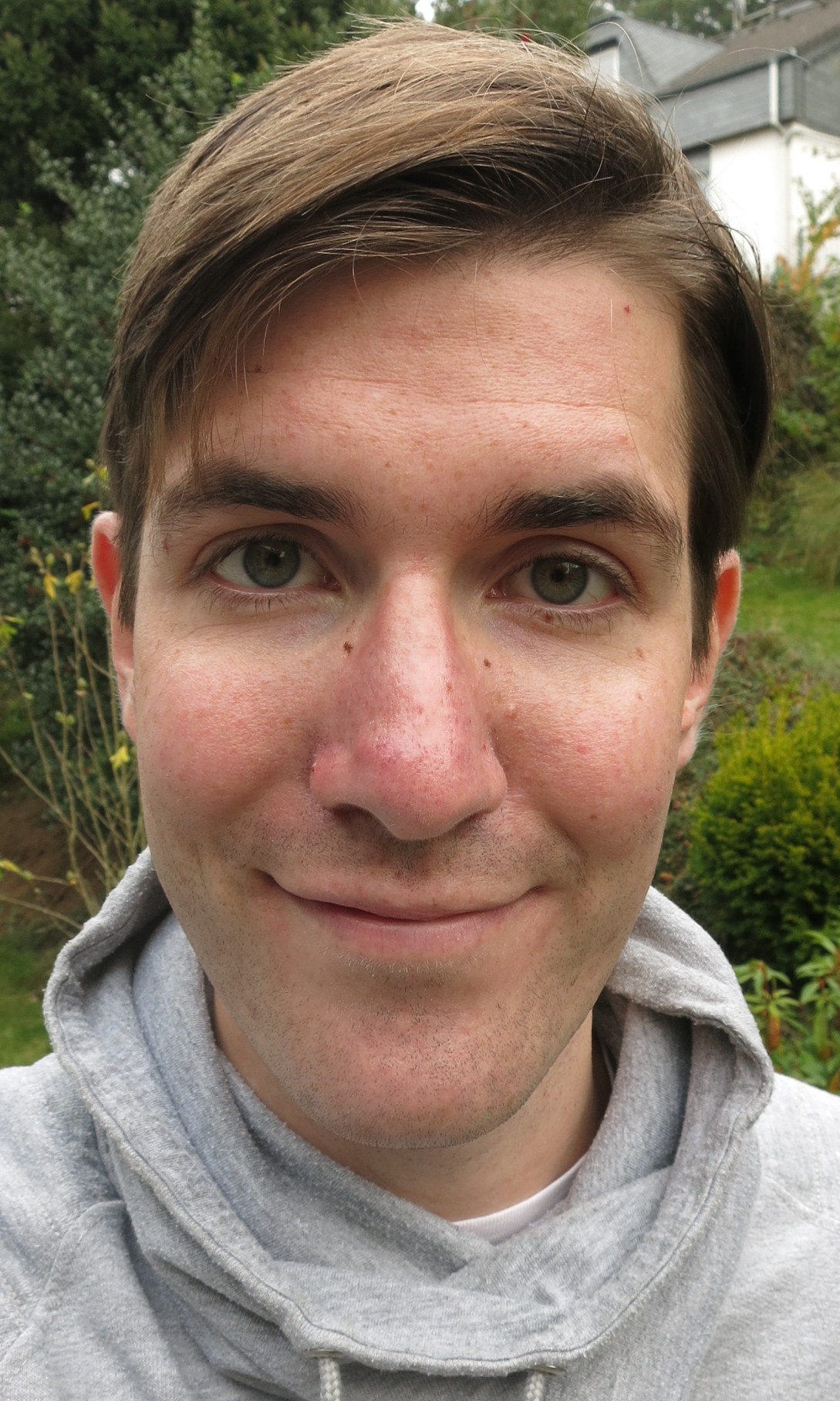}}]{Tobias Willig}
received a BASc degree in Computer Science from Bonn-Rhein-Sieg University of Applied Sciences in 2011 with a grade of 1.3, and a MSc degree in Computer Science from University of Bonn with an excellent grade of 1.2. He finished his Master's degree in 2015 with a thesis on temporal segmentation
of human motion capture data.
\end{IEEEbiography}


\begin{IEEEbiography}[{\includegraphics[width=1in,height=1.25in,clip,keepaspectratio]{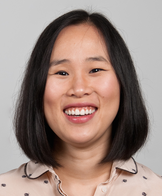}}]{Angela Yao}
is currently an Assistant Professor at the Institute of Computer Science at the University of Bonn. She received a BASc degree in Engineering Science from the University of Toronto in 2006 and a PhD in Information Technology and Electrical Engineering from ETH Zurich in 2012. Her research interests are in computer vision and machine learning, with special focus on human pose estimation and action recognition.
\end{IEEEbiography}

\begin{IEEEbiography}[{\includegraphics[width=1in,height=1.25in,clip,keepaspectratio]{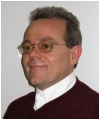}}]{Reinhard Klein}
studied Mathematics and Physics at the University of T\"ubingen, Germany, from where he received his MS in Mathematics (Dipl.-Math.) in 1989 and his PhD in computer science in 1995. In 1999 he received an appointment as lecturer ("Habilitation") in computer science also from the University of T\"ubingen, with a thesis in computer graphics. In September 1999 he became an Associate Professor at the University of Darmstadt, Germany and head of the research group Animation and Image Communication at the Fraunhofer Institute for Computer Graphics. Since October 2000 he is professor at the University of Bonn and director of the Institute of Computer Science II.
\end{IEEEbiography}

\begin{IEEEbiography}[{\includegraphics[width=1in,height=1.25in,clip,keepaspectratio]{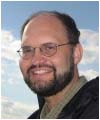}}]{Andreas Weber}
studied mathematics and computer science at the
Universities of T\"ubingen, Germany and Boulder, Colorado, U.S.A. From the
University of T\"ubingen he received his MS in Mathematics (Dipl.-Math) in
1990 and his PhD (Dr. rer. nat.) in computer science in 1993. From 1995
to 1997 he was working with a scholarship from Deutsche
Forschungsgemeinschaft as a postdoctoral fellow at the Computer Science
Department of Cornell University. From 1997 to 1999 he was a member of
the Symbolic Computation Group at the University of T\"ubingen, Germany.
From 1999 to 2001 he was a member of the research group Animation and
Image Communication at the Fraunhofer Institut for Computer Graphics.
\end{IEEEbiography}





\appendices
\section{Data Source Documentation}\label{datasources}

\begin{figure}[htbp]
\centering
\subfloat{\includegraphics[width=0.45\textwidth]{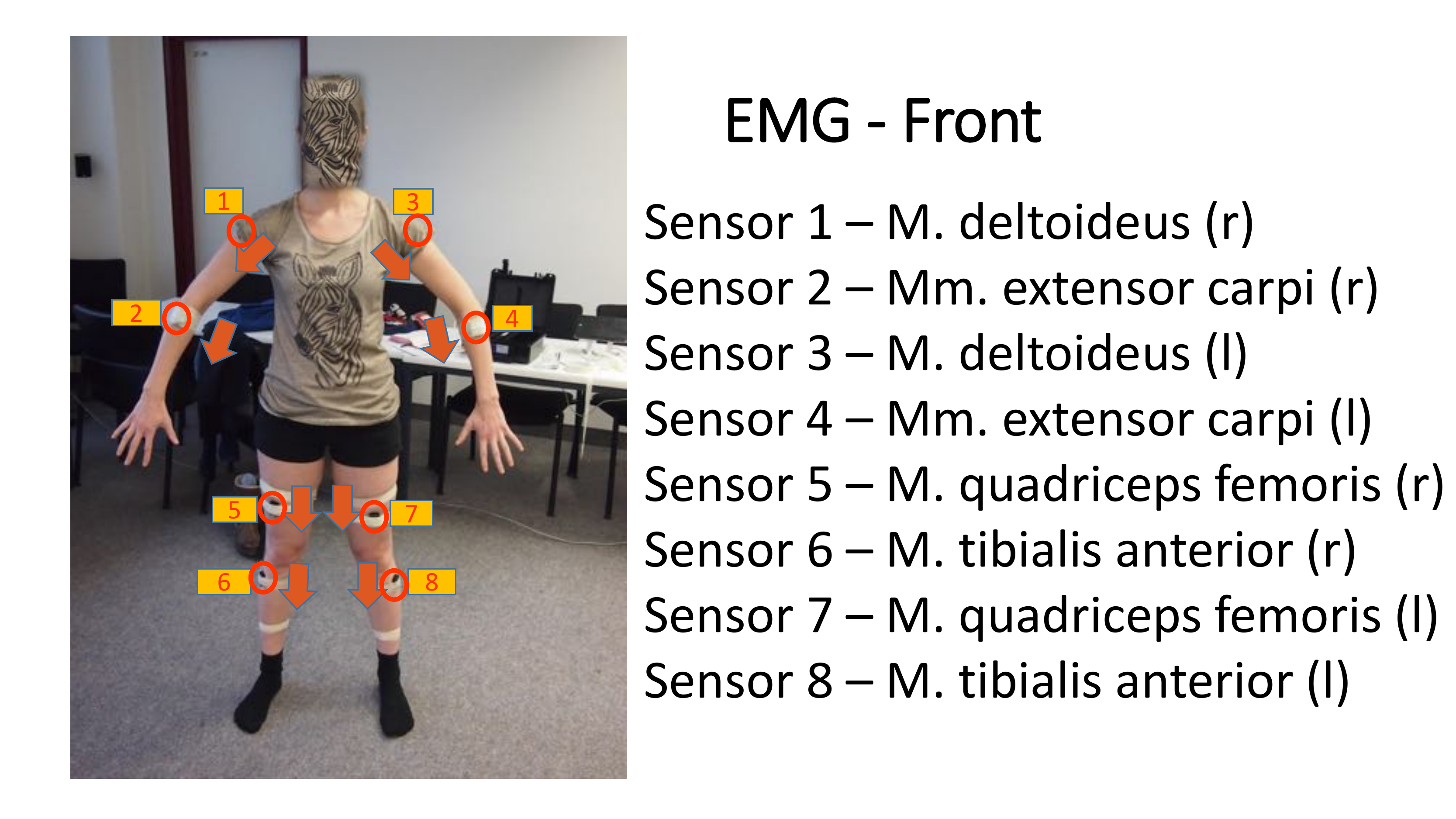}}\vfill
\subfloat{\includegraphics[width=0.45\textwidth]{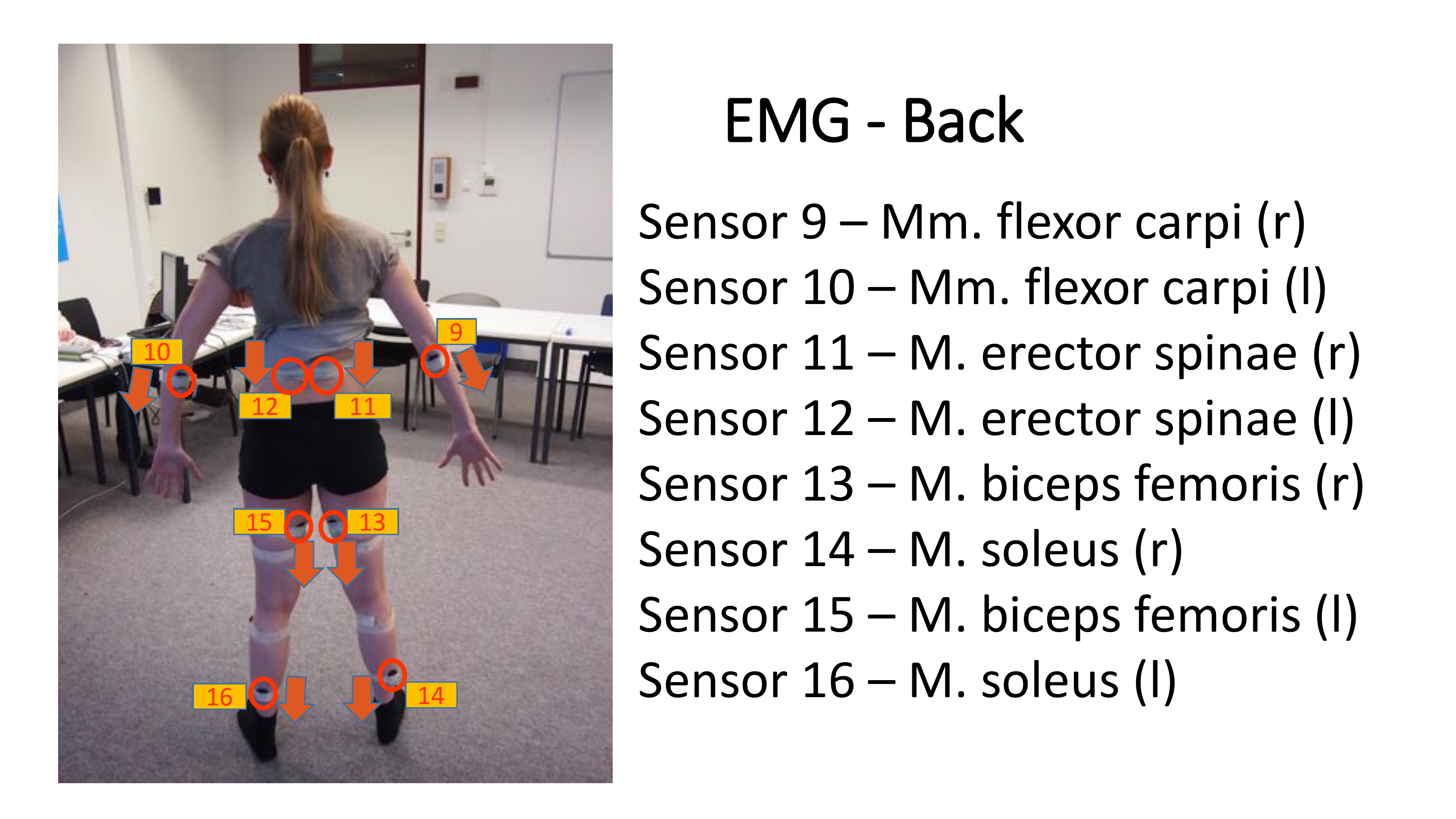}}
\caption{Documentation of EMG sensor placement on a human subject seen from the front and back, respectively. Orange arrows point to direction of $y$-axis.}
\label{fig:sensor_documentation}
\end{figure}

We recorded our EMG and acceleration data using a combined wireless Delsys Trigno system. The sensor setup consisted of $16$ sensors placed on larger muscle groups of the trial subject (see Figure~\ref{fig:sensor_documentation} for a list of locations and a visualization of the hardware attachment). Both sensor attachment and recordings were supervised by an accredited expert.

A selection of trials performed originally by subject $86$ of the CMU database was re-enacted by another human subject as true to original as possible. This was done to compare the performance of our segmentation method on other data modalities while maintaining control of the activities represented by the data streams. There are some minor deviations from the original scripts due to different geometry of the location (e.g. in one of our trials, originally $86\_12$, the subject had to walk some additional steps after sweeping the floor and before reaching the white board). The total length of the trials differs from the original slightly due to similar reasons. Nevertheless, the the sequences contain the same activity classes as the original sequences, and have mostly the same number of repetitions and are therefore suitable for a qualitative comparison.
%
%
%

\section{Symmetry Types}\label{app:symmetry}
Schematic representations of self-similarity matrices for each of the possible cases, discussed in Section~\ref{motion_symmetry} are given in Table~\ref{tab:symmetries}.
For symmetrical motions, there is no difference in the diagonal structure in the SSSM from the original and mirrored features. This leads to no additional cuts when symmetry is exploited for the segmentation.
Phase-shifted symmetry can occur with and without speed variation. Without any speed variations, a more diverse structure in the self-similarity matrices occurs in both the original and mirrored features. For the original features, more block-shaped parts may appear on each diagonal, while in the mirrored setting, all diagonals aggregate to a more curvy pattern.
In the asynchronous case, the mirrored self-similarity matrix shows no diagonal structures at all, adding to no additional cuts. 
When symmetries are mixed, the SSSMs are characterized locally according to the corresponding symmetry type.

\begin{table}[h!]
\caption{Schematic self similarity matrices for various types of symmetries.}
\label{tab:symmetries}
\centering

\begin{tabular}{cl}
\hline
    \begin{minipage}{0.2\textwidth}
        \includegraphics[width=\textwidth]{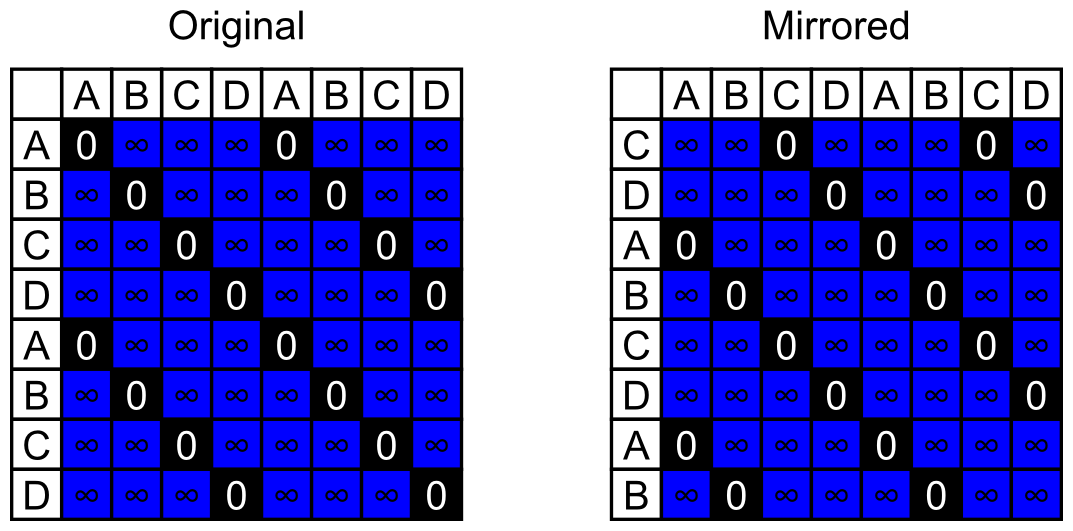}
        \includegraphics[width=\textwidth]{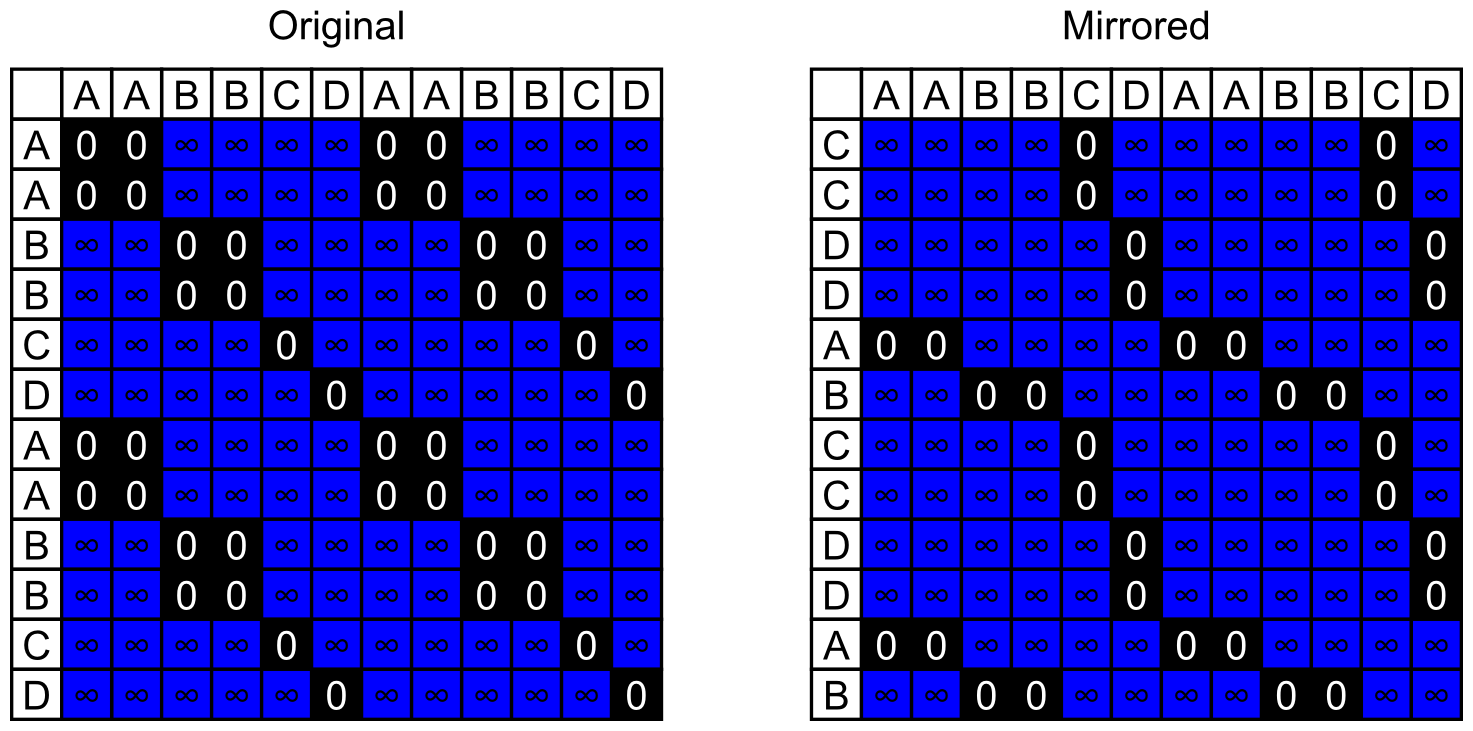}
    \end{minipage}
&
    \begin{minipage}{0.25\textwidth}
        \vspace{1mm}
        Two cases of phase-shifted symmetries. The first matrix shows that the original and mirrored features are disjoint, i.e. the diagonal structures in the SSSM are at different locations.  The second is a similar situation where the original motion shows some speed variation, leading to a more diverse structure in both matrices. The mirrored features match the originals in some sense but the variation is seen in the slopes of the diagonals.
        \vspace{1mm}
    \end{minipage}
\\
\hline
    \begin{minipage}{0.2\textwidth}
        \includegraphics[width=\textwidth]{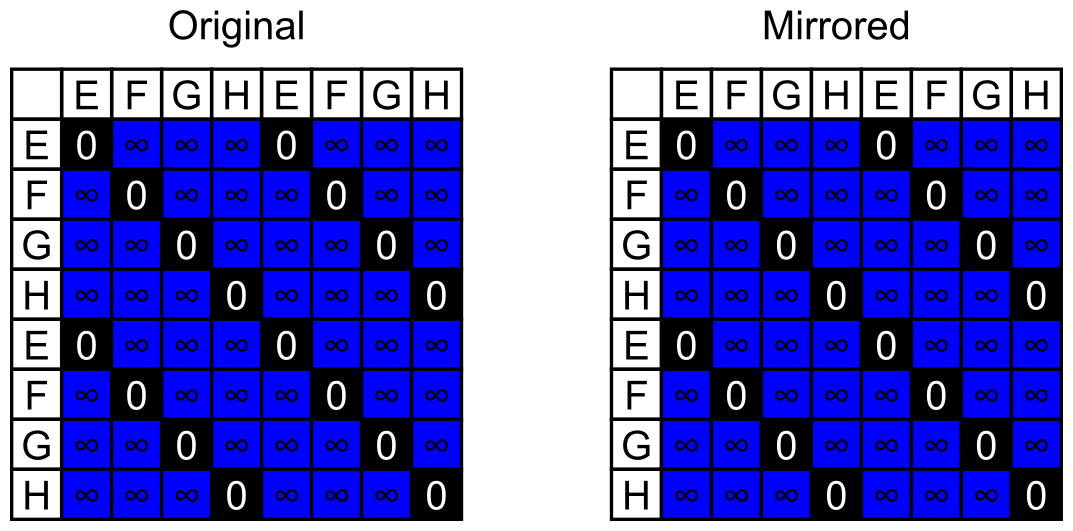}
    \end{minipage}
&
    \begin{minipage}{0.25\textwidth}
    \vspace{1mm}
        Example of a symmetric motion: The diagonal structures in the SSSM are at exactly the same locations, leading to no additional cuts.
    \vspace{1mm}
    \end{minipage}
\\
\hline
    \begin{minipage}{0.2\textwidth}
        \includegraphics[width=\textwidth]{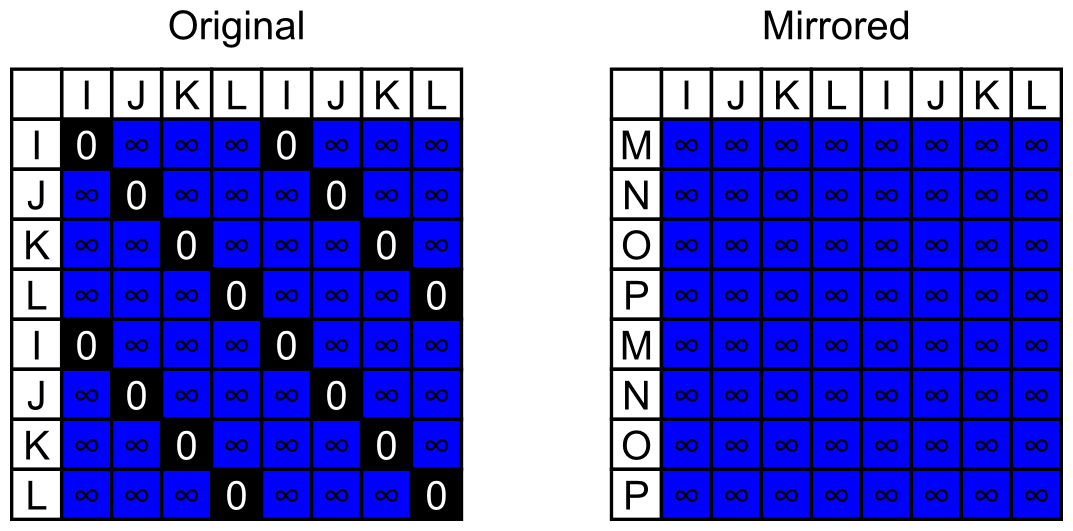}
    \end{minipage}
&
    \begin{minipage}{0.25\textwidth}
    \vspace{1mm}
        Example of an asynchronous motion: The diagonal structures in the original SSSM have no matches in the mirrored version, again leading to no additional cuts.
    \vspace{1mm}
    \end{minipage}
\\
\hline
    \begin{minipage}{0.2\textwidth}
        \includegraphics[width=\textwidth]{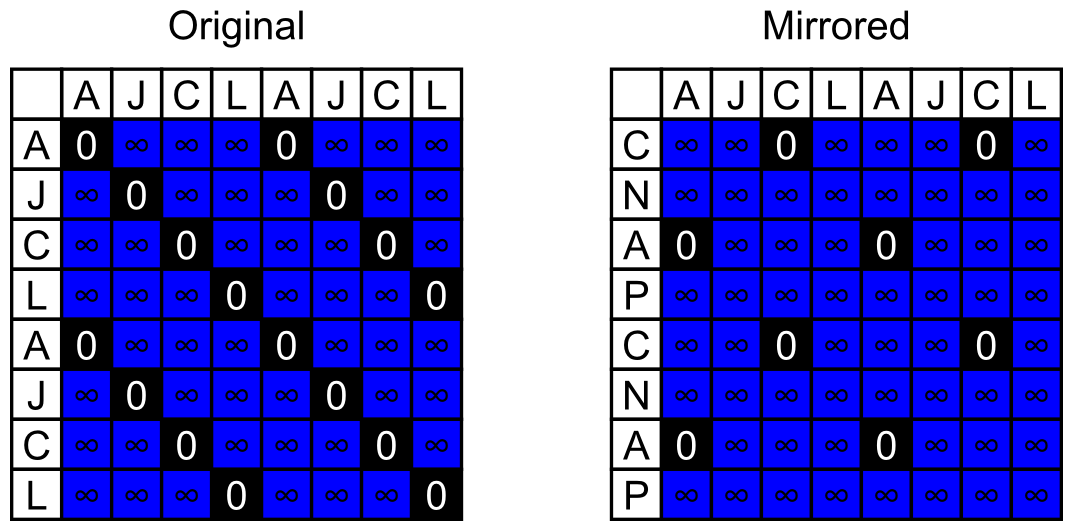}
    \end{minipage}
&
    \begin{minipage}{0.25\textwidth}
    \vspace{1mm}
        Example of a motion with different symmetries, with local matches for the original diagonals in the mirrored version. This means that the motion has phase-shifted symmetrical phases interrupted by other phases which are asynchronous.
    \vspace{1mm}
    \end{minipage}
\\
\hline

\end{tabular}

\end{table}



\ifCLASSOPTIONcompsoc
  \section*{Acknowledgments}
\else
  \section*{Acknowledgment}
\fi

The authors would like to thank Kristian Welle (Bonn University Hospital, Clinic for Orthopaedics and Trauma Surgery) for supporting with the EMG measurements.
This work was partially supported by Deutsche Forschungsgemeinschaft (DFG) under research grants KR 4309/2-1.

\ifCLASSOPTIONcaptionsoff
  \newpage
\fi

\end{document}